\documentclass[manuscript,screen,sigconf]{acmart}
\AtBeginDocument{%
  \providecommand\BibTeX{{%
    \normalfont B\kern-0.5em{\scshape i\kern-0.25em b}\kern-0.8em\TeX}}}

\renewcommand\footnotetextcopyrightpermission[1]{}

\acmConference[]{}{}{}

\usepackage{setspace} 

\usepackage{etoolbox}
\AtBeginEnvironment{quote}{\par\singlespacing\small}

\usepackage{balance}

\newcommand{\qt}[1]{``#1''}

\newcommand{\modelname}{{Our Model}}

\newcommand{\xhdr}[1]{\vspace{1.7mm}\noindent{{\bf #1.}}}

\graphicspath{{./fig/}}

\begin{document}

\title[Self-supervised Pretraining and Transfer Learning Enable Flu and COVID-19 Predictions in Small Mobile Sensing Datasets]{Self-supervised Pretraining and Transfer Learning Enable Flu and COVID-19 Predictions in Small Mobile Sensing Datasets}



\author{Mike A. Merrill}
\email{mikeam@cs.washington.edu}
\affiliation{%
  \institution{University of Washington}
  \city{Seattle}
  \state{Washington}
  \country{USA}
}

\author{Tim Althoff}
\email{althoff@cs.washington.edu}
\affiliation{%
  \institution{University of Washington}
  \city{Seattle}
  \state{Washington}
  \country{USA}
}

\begin{abstract}
Detailed mobile sensing data from phones, watches, and fitness trackers offer an unparalleled opportunity to quantify and act upon previously  unmeasurable behavioral changes in order to improve individual health and accelerate responses to emerging diseases. 
Unlike in natural language processing and computer vision, deep representation learning has yet to broadly impact this domain, in which the vast majority of research and clinical applications still rely on manually defined features and boosted tree models or even forgo predictive modeling altogether due to insufficient accuracy.
This is due to unique challenges in the behavioral health domain, including
very small datasets ($\sim \!\! 10^1$ participants), which frequently contain missing data, consist of long time series with critical long-range dependencies (length>$10^4$), and extreme class imbalances (>$10^3$:1).

Here, we introduce a neural architecture for multivariate time series classification designed to address these unique domain challenges.
Our proposed behavioral representation learning approach combines novel tasks for self-supervised pretraining and transfer learning to address data scarcity, 
and captures long-range dependencies across long-history time series through transformer self-attention following convolutional neural network-based dimensionality reduction. 
We propose an evaluation framework aimed at reflecting expected real-world performance in plausible deployment scenarios.
Concretely, we demonstrate (1) performance improvements over baselines of up to 0.15 ROC AUC across five prediction tasks,
(2) transfer learning-induced performance improvements of 16\% PR AUC in small data scenarios,
and (3) the potential of transfer learning in novel disease scenarios through an exploratory case study of zero-shot COVID-19 prediction in an independent data set. 
Finally, we discuss potential implications for medical surveillance testing.






\end{abstract}

\maketitle

\newcommand{\modelfigure}{
\begin{figure*}
  \includegraphics[width=\textwidth]{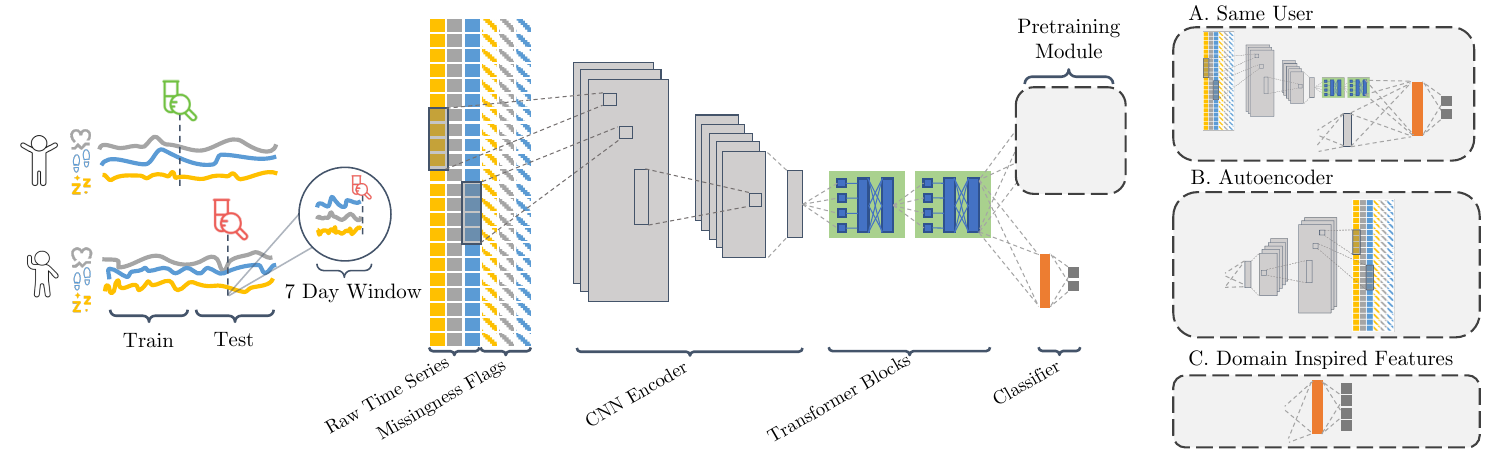}
  \caption{Our Model (Section~\ref{subsec:model}) combines a CNN encoder for learning hierarchical  and temporal features from raw time series data and a transformer for learning long-range relationships between these features. Additionally, we provide three novel self-supervised pretraining tasks for learning from unlabeled data (Section~\ref{subsec:self_supervised_tasks}).}
  \Description{}
  \label{fig:teaser}
\end{figure*}

}

\newcommand{\mainTaskTable}{
    \begin{table*}
        

        \begin{tabular}{lllp{1in}p{1in}p{1in}p{1in}}
        \toprule
                           &               & &  XGBoost &    CNN &  CNN-Transformer        &  Our Model \\
                      Task & Class Balance & Metric &          &        &                  &            \\
        \midrule
                    Flu Positivity & 1:2,760 & PR AUC &    0.003 &  0.007 &            0.007    &      \textbf{0.011 } \\
                                              && ROC AUC &    0.708 &  0.860 &            0.884 &      \textbf{0.887 } \\
                    Severe Fever & 1:643 & PR AUC &    0.013 &  0.048 &            0.039        &      \textbf{0.056 } \\
                                            &  & ROC AUC &    0.741 &  0.801 &            0.790 &      \textbf{0.818 } \\
                    Severe Cough & 1:132 & PR AUC &    0.018 &  0.017 &            0.023        &      \textbf{0.023 } \\
                                              && ROC AUC &    0.704 &  0.690 &            0.697 &      \textbf{0.708 } \\
                    Severe Fatigue & 1:78 & PR AUC &    0.032 &  0.038 &            0.038       &      \textbf{0.074*} \\
                                             & & ROC AUC &    0.708 &  0.699 &            0.713 &      \textbf{0.758*} \\
                    Flu Symptoms &  1:37 & PR AUC &    0.044 &  0.032 &            0.042        &      \textbf{0.066*} \\
                                             & & ROC AUC &    0.647 &  0.612 &            0.640 &      \textbf{0.671*} \\
        \bottomrule
\end{tabular}

    \caption{Results on all tasks for our model. *Indicates $p<0.05$ (Delong). Note that while substantial class imbalance precludes statistically significant results on some tasks (\qt{Flu Positivity}, \qt{Severe Fever}, and \qt{Severe Cough}), \modelname~performs better than all baselines and ablations when jointly evaluating performance across \emph{all} tasks to increase statistical power (Figure~\ref{fig:CritDiff}).}
    \label{tab:mainTasks}
\end{table*}}

\newcommand{\tableTaskFeatures}{
\begin{table}[h]
    \centering
    \begin{tabular}{ll}
         & \\
        \toprule
        Resting HR & Avg. heart rate (HR) while not moving\\
        Main Minutes in Bed & Longest duration of minutes in bed\\
        Sleep Efficiency & Time spent sleeping over time in bed\\
        Nap Count & Number of naps\\
        Total Asleep Minutes & Total time spent sleeping\\
        Total in Bed Minutes & Total time spent in bed\\
        Active Calories & Calories burned from exercise\\
        Calories Out & Total calories burned\\
        Base Metabolic Rate & Calories passively burned\\
        Sedentary Minutes & Time spent not moving\\
        Lightly active minutes & Time spent lightly active\\
        Fairly active minutes & Time spent lightly exercising\\
        Very active minutes & Time spent actively exercising\\
        Missing HR & Indicator for missing HR data\\
        Missing Sleep & Indicator for missing sleep data\\
        Missing Steps & Indicator for missing steps\\
        Missing Day & Indicator for missing all data\\
        \hline
    \end{tabular}
    \caption{Summary of manually defined features, calculated for every user and on each day. \qt{Missing} features are binary variables which are 1 if more than one hour of data is missing, and 0 otherwise.}
    \label{tab:features}
    \vspace{-20pt}
\end{table}
}

\newcommand{\figFluTwentyFold}{
\begin{figure}
    \includegraphics[width=0.95\columnwidth]{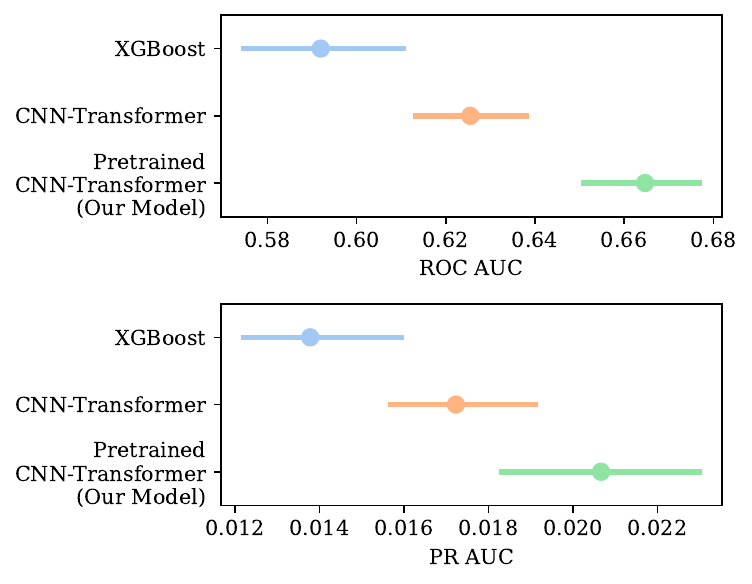}
    \vspace{-5mm}
    \caption{
    Comparison of XGBoost, CNN-Transformer, and a pretrained CNN-Transformer (Our Model) on the \qt{Flu Positivity} task (Section~\ref{subsec:pred_tasks}) using training data from ten participants. Neural models outperform XGBoost, and pretraining yields additional improvement. This indicates that pretraining helps models generalize to tasks with limited training data. Horizontal lines represent 95\% confidence intervals. Differences in means are significantly significant  (all $p\leq0.05$; Mann-Whitney U).}
    \label{fig:flu_twenty_fold}
\end{figure}
}

\newcommand{\figPreTraining}{
\begin{figure}
    \includegraphics[width=\columnwidth]{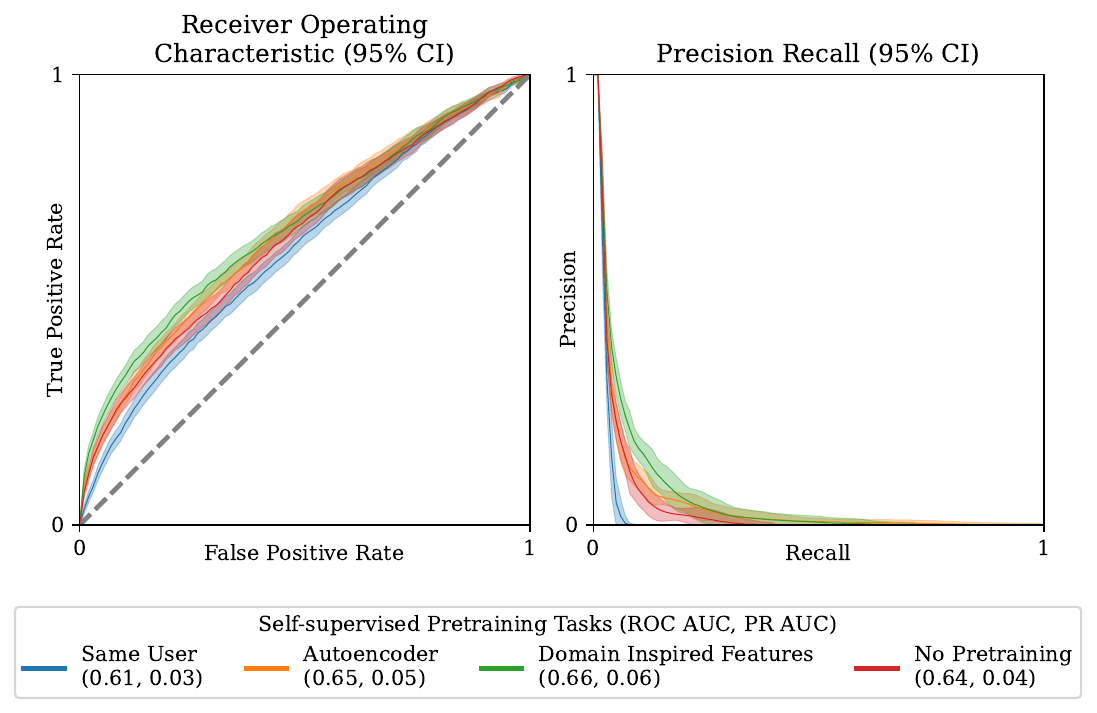}
    \caption{Comparison of self-supervised pretraining tasks (Section~\ref{subsec:self_supervised_tasks}) on the \qt{Flu Symptoms} task. The \qt{Domain Inspired Features} task, which integrates domain knowledge, performs best relative to a non-pretrained model.}
    \label{fig:pretrainingResults}
\end{figure}
}

\newcommand{\figCritDiff}{
\begin{figure}
    \includegraphics[width=0.95\columnwidth]{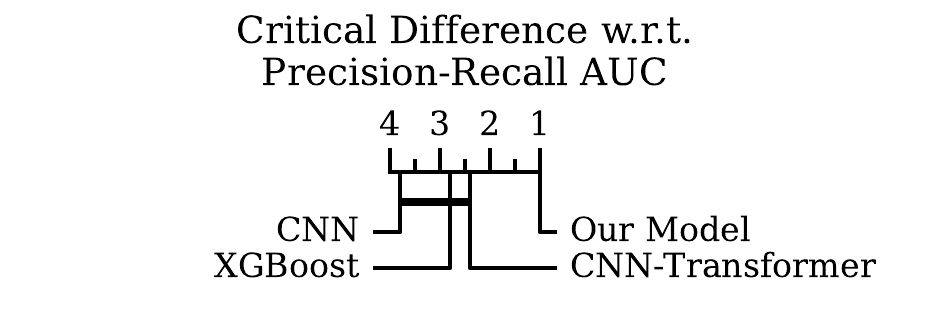}
    \vspace{-6mm}
    \caption{Critical Difference Plot \cite{carbonell_comparison_2000} between models at $\alpha=0.1$. Numbers indicate each model's average ranking on the single domain prediction tasks (Section~\ref{subsec:pred_tasks}), while the thick dark line connects models which are not significantly different from one another. This demonstrates that Our Model, which uses pretraining and models missing data, significantly outperforms CNNs, XGBoost, and CNN-Transformers across tasks (average rank=1.0).}
    \label{fig:CritDiff}
\end{figure}
}

\newcommand{\binnedMissing}{
\begin{figure}
    \includegraphics[width=0.4\columnwidth]{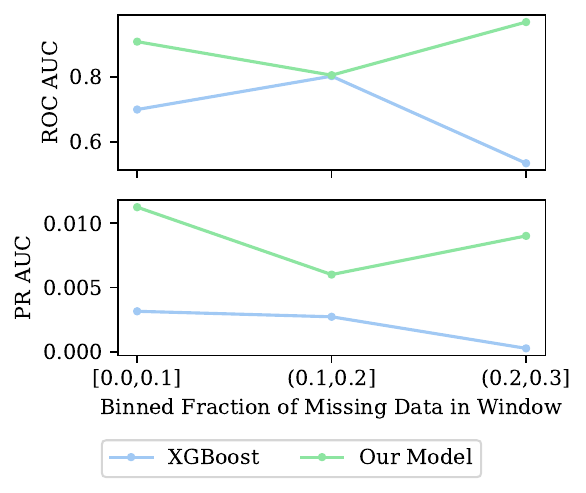}
    \caption{Performance on the \qt{Flu Positivity} task for binned levels of missing data. Missingness is defined as the fraction of minutes with heart rate data over the duration of the accompanying seven day window. 90\% of labels have less that 30\% missingness, making positive labels sparse, and so we do not compare methods past this threshold.}
    \label{fig:binnedMissing}
\end{figure}
}

\newcommand{\pretrainConcept}{
\begin{figure}
    \includegraphics[width=0.8\columnwidth]{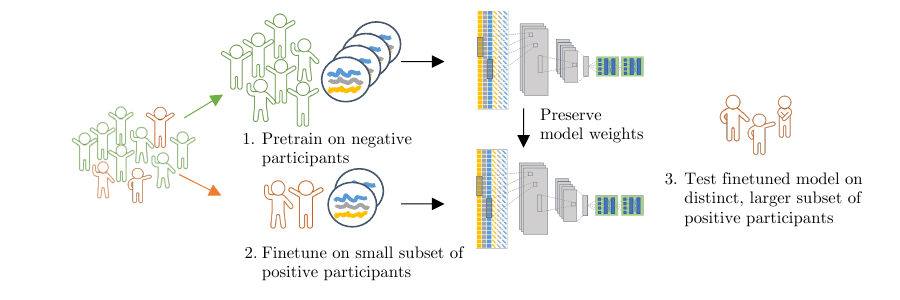}
    \caption{Overview of the cross validation split for our simulation of small datasets (Section~\ref{subsec:sim_study})}
    \label{fig:pretrainedConcept}
\end{figure}
}

\newcommand{\domainTransferTable}{
    \begin{table}
        \begin{tabular}{lrr}
                      & XGBoost & Our Model\\
        \hline                      
        Zero-shot PR AUC        & 0.005   &  0.018\\
        Zero-shot ROC AUC       & 0.51    &  0.68\\
        \bottomrule
    \end{tabular}
    \caption{Performance on zero-shot COVID-19 Prediction (Section~\ref{subsec:domain_transfer}). Our model's superior performance indicates that CNN-Transformers pretrained on the \qt{Domain Inspired Features} task (Section~\ref{subsec:self_supervised_tasks}) learn generalizable features which transfer to new domains.}
    \label{tab:domainTransfer}
    \vspace{-20pt}
    \end{table}
}

\newcommand{\tableDataStats}{
    \begin{table}
        \begin{tabular}{lr}
            \textbf{Feature} & \textbf{Description}\\
            \toprule
            Number of participants & 5196\\
            Average number of days of data & 114\\
            Portion of days containing missing data & 93\% \\
            Mean age ($\pm$SD) & 37.7 (10.2)\\
            Portion Female & 72\% \\
            Mean BMI ($\pm$SD) & 30.3 (20.3)\\
            Number of US States Represented & 50\\
            \bottomrule
        \end{tabular}
        \caption{Summary statistics for the Homekit Flu Monitoring Study (Section~\ref{subsec:dataset})}
        \label{tab:data_stats}
    \end{table}
}

\newcommand{\tableZeroShotFeatures}{
\begin{table}[h]
    \centering
    \begin{tabular}{ll}
        \textbf{Feature} & \textbf{Description}\\
        \toprule
        Resting HR 95\textsuperscript{th} Pct & 95\textsuperscript{th} percentile of resting HR\\
        Resting HR 50\textsuperscript{th} Pct & 50\textsuperscript{th} percentile of resting HR\\
        Resting HR std. & Standard deviation of resting HR\\
        Awake HR 95\textsuperscript{th} Pct & 95\textsuperscript{th} percentile of HR while awake\\
        Steps Streak 95\textsuperscript{th} Pct & 95\textsuperscript{th} percentile of continuous steps\\
        Steps Streak 50\textsuperscript{th} Pct & 50\textsuperscript{th} percentile of continuous steps\\
        Total Minutes in Bed & Number of minutes spent in bed during duration\\
        Sleep Minutes & Number of minutes spent asleep during duration\\
        Total Steps & Total Number of steps during duration\\
        Missing HR & Indicator for missing HR data\\
        Missing Sleep & Indicator for missing sleep data\\
        Missing Steps & Indicator for missing steps\\
        Missing Day & Indicator for missing all data\\
        \hline
    \end{tabular}
    \caption{Summary of manually defined features used for XGBoost baseline in the zero shot experiment (Section~\ref{subsec:domain_transfer}) calculated for every user and on each day. \qt{Missing} features are binary variables which are 1 if more than one hour of data is missing, and 0 otherwise.}
    \label{tab:zero_shot_features}
\end{table}
}
\section{Introduction}


Mobile sensing data from phones, watches, and fitness trackers offer an unparalleled opportunity to track complex behavioral changes and symptoms, detect high risk individuals in large populations, and deploy targeted interventions.
Because many conditions manifest themselves through behavioral and physiological changes (e.g., reduced activity, disrupted sleep, increased heart rate),
leveraging these data 
could minimize the impact of emerging diseases.
Currently, such conditions exact a massive toll (e.g.,~\cite{mezlini2021estimating}), with contagious respiratory illnesses such as COVID-19 (or influenza/flu) rising to the second leading cause of death in the U.S. in January 2022.

Despite the enormous potential and availability of these data for well over a decade, broad and tangible impacts on population health have yet to be realized. 
For example, consider their limited impact on COVID-19, which reduced gross global product by \$28 trillion~\cite{imf2020covid}; except for contact tracing apps, which do not require predictive modeling, the global COVID-19 response made no significant use of these data
beyond research studies.

While neural representation learning approaches have provided transformative performance improvements across Natural Language Processing (NLP) and Computer Vision (CV) ~\cite{mikolov2013distributed,devlin2019bert,lewis2019bart,lan2019albert,liu2019roberta,he2016resnet,krizhevsky2012alexnet},  currently these techniques are rarely adopted for mobile sensing research and applications.
In contrast to typical NLP and CV benchmark datasets, mobile sensing data are usually very limited in size (often less than 20 individuals due to arduous and expensive data collection~\cite{xu2021understanding}), frequently contain missing data (93\% of days in our data; e.g. when device is not used or charging), consist of long history time series (>10,000 min-by-min time steps for one week of data) with relevant long range dependencies (e.g. changes in heart rate across multiple days), and feature extreme class imbalances (up to 2760:1 in our data; e.g. because most people are not sick on most days)~\cite{xu2021understanding}.
Therefore, researchers have often been limited to using small datasets and less data hungry non-neural models such as boosted tree models with hand-crafted features which typically perform worse (e.g.,~\cite{xu2021understanding,laport-lopez_review_2020,zhang_passive_2021, nair_understanding_2019, lin_healthwalks_2020, hafiz_wearable_2020, buda_outliers_2021, mairittha_crowdact_2021, meegahapola_one_2021}).

In this paper, we introduce a neural architecture for multivariate time series classification specifically designed for these unique domain challenges.
Specifically, 
(1) this model learns directly from raw minute-level sensor data (in contrast to prevailing use of manually defined features; Section~\ref{subsec:model}),
(2) leverages novel self-supervised pretraining tasks (Section~\ref{subsec:self_supervised_tasks}) and transfer learning to improve performance in datasets of limited size without requiring additional supervision, 
(3) directly models potentially informative missingness patterns instead of excluding participants with missing data,
(4) captures long-range dependencies across long-history time series through transformer layers~\cite{vaswani2017attention}, 
while (5) reducing the input sequence length to these transformer layers through hierarchical feature extraction of convolutional neural networks (CNNs).\footnote{Because the full self-attention mechanism of transformers has computational and memory requirements that are quadratic with the input sequence length \cite{beltagy_longformer_2020}}

Next, we present a framework of best practices for evaluating mobile sensing models (Section~\ref{sec:evaluation}), which describes (1) how to avoid massively overestimating model performance relative to expected real-world performance, and (2) an approach to reduces the statistical uncertainty introduced by inherent extreme class imbalances (up to 1:2,760 in our evaluation data; Section~\ref{subsec:dataset}) that is based on jointly comparing model performance across multiple prediction tasks. We then apply this framework to the evaluation of the proposed model across four experiments.


In {\sc Experiment 1}~(Section~\ref{subsec:pred_tasks}) we evaluate model performance across five single domain prediction tasks related to predicting the flu with FitBit wearable data, and show that CNN encoders, transformer blocks, modeling missingness, and self-supervised pretraining significantly increase predictive performance, up to 0.15 ROC AUC relative to common baselines.

In {\sc Experiment 2}~(Section \ref{subsec:results_pretraining}) we compare three novel self-supervised pretraining tasks and show that a task which incorporates basic domain expertise performs best.

In {\sc Experiment 3}~(Section \ref{subsec:results_transfer_learning}) we demonstrate transfer learning of pretrained behavioral representations. Specifically, we simulate 20 separate small data studies with only ten participants each for training. We show that finetuning a pretrained model (trained on a separate self-supervision task on an independent set of participants) on these ten participants outperforms training from scratch with a $\sim16\%$ improvement in precision-recall AUC.

In {\sc Experiment 4}~(Section \ref{subsec:domain_transfer}) we extend the previous transfer learning setting to an exploratory case study of zero-shot COVID-19 prediction in a small third party dataset. In this zero-shot paradigm, without any training on COVID-19 cases, the proposed pretrained model achieves a 0.62 ROC AUC, while an XGBoost baseline cannot exceed near-random performance (0.51 ROC AUC).\footnote{Note that statistical power is limited due to the small dataset size, and therefore we first include the repeated simulation study of {\sc Experiment 3} to demonstrate robustness of transfer learning performance.} This demonstrates that the pretraining of the proposed model architecture is able to learn generalizable features that enable significant performance improvements across multiple domains (flu and COVID-19). 


Finally, we reflect on these advances and potential implications in the the context of the medical literature on surveillance testing (Section ~\ref{sec:discussion}).
Advances in model performance especially on small datasets and novel disease scenarios could support a more widespread use of mobile sensing data as well as enable rapid deployments in emerging disease scenarios. For example, in the crucial early days of the COVID-19 pandemic, laboratory testing was not widely available, and many positive cases remained undetected. In this setting, a generalizable pretrained model could be fine-tuned on the few test cases already available, and used to identify members of a population who may be infected and should be targeted for additional testing or interventions~\cite{brook_optimizing_2021,nestor2021dear,quer_wearable_2020}. 
It is promising to note that the proposed model enables predictive performance comparable to some flu and COVID-19 rapid antigen tests (ca. 0.68 to 0.88 ROC AUC~\cite{bachman_clinical_2021,chu_performance_2012}). 
Still, we emphasize that additional research and validation experiments are needed to support the use of predictive models such as ours in public health strategy and policy.

In summary, our contributions include:
\begin{itemize}
    \item A neural architecture for multivariate time series classification in mobile sensing and novel set of pretraining tasks (Section~\ref{sec:methods})
    \item A framework for evaluating mobile sensing models, which provides best-practices for selecting realistic prediction tasks and mitigating inherent statistical uncertainty during model selection (Section~\ref{sec:evaluation})
    \item An empirical evaluation demonstrating that the proposed approach significantly improves prediction performance on small datasets through transfer learning in both flu predictions and a case study of a novel zero-shot COVID-19 prediction task (Section~\ref{sec:results}).
\end{itemize}

We make our model publicly available for use by researchers and practitioners at \url{https://github.com/behavioral-data/FluStudy}, so that they may use it as an initialization for their own prediction tasks. 



\section{Related Work}

Our model builds upon prior work in neural methods and transfer learning for behavioral sensing and modeling. Our model is the first to learn generalizable feature representations from long-history multivariate time series to enable transfer learning in small datasets.

\subsection{Neural Models for Time Series Classification in Mobile Sensing}
Behavioral data has been modeled and mined using deep learning techniques across a variety of domains, including human activity recognition (using CNN)~\cite{yao2017deepsense},
personalized fitness recommendation (using stacked LSTM~\cite{hochreiter1997long})~\cite{ni2019modeling},
mood prediction (using RNN, GRU, or autoencoder)~\cite{suhara2017deepmood,cao2017deepmood,spathis2019sequence},
stress prediction (using LSTM and autoencoder)~\cite{li2020extraction},
health status prediction (using CNN and cross-attention)~\cite{hallgrimsson2018learning}, 
and personality prediction~\cite{wu2020representation}. Two studies experimented with multi-head attention and convolution as we do here, but neither paper applies this architecture to transfer learning \cite{song2018attend,tang2021selfhar}. Liu et. al \cite{liu_fitbeat_2022} apply a CNN autoencoder to raw sensor data to predict COVID-19, but do not experiment with transformer layers nor transfer learning as we do here.

Until very recently, the state of the art in time series classification has eschewed deep learning in favor of more traditional statistical learning methods \cite{fawaz_deep_2019}. Fawaz et. al \cite{fawaz_deep_2019} propose a set of benchmark datasets and tasks for time series classification, but relative to the minute-level time series we model here these datasets are shorter (at most 2,000 observations in length), do not contain missing data, and are mostly univariate.

\subsection{Self-Supervised Learning in Behavioral Modeling}
In the broader field of self-supervision for time series classification, the most relevant work is Zerveas et. al \cite{zerveas_transformer-based_2021}, who use a simple linear projection to shrink the input multivariate time series to the scale supported by transformers. Zhang et. al~\cite{zhang_deep_2019} use an LSTM to learn self-supervised representations of multivariate timeseries, but apply this method to anomaly detection. Transfer learning remains a \qt{grand challenge} for mobile sensing \cite{wang2019deep}, and has been explored in human activity recognition~\cite{ma2020transfer}, stress and mood prediction~\cite{jaques2017multitask,li2020extraction}, and forecasting adverse surgical outcomes in an ICU~\cite{chen2020forecasting}. Hallgrímsson et. al~\cite{hallgrimsson2018learning} use a CNN autoencoder to forecast heart rate from steps and sleep data, but do not predict acute events like viral infection. Kolbeinsson et. al~\cite{kolbeinsson_self-supervision_2021} pretrain transformers to predict one of several handcrafted features on a day given the previous day's aggregated sensor data, but do not model minute-level raw time series as we do here. However, none of these applications focus explicitly on model performance on small datasets. Tang et. al \cite{tang2021selfhar} focuses on performance on small datasets, but unlike our work does not evaluate with a zero-shot, out-of-domain task.
\section{Methods}
\label{sec:methods}
Here, we detail our model which is composed of a CNN encoder for learning hierarchical and temporal features, effectively reducing the dimensionality, and a transformer for learning potentially long-range relationships between these features. Then, we motivate and describe a set of self-supervised pretraining tasks that can be used to boost overall model performance.

\modelfigure

\subsection{Our Model}
\label{subsec:model}
\modelname~is composed of a convolutional encoder, a stack of transformer blocks, and a final densely connected linear layer that is used for classification. Intuitively, the convolutional encoder learns a compressed, hierarchical feature representation of the raw time series data, while the transformer learns relationships between these features. An overview of our model architecture is given in Figure \ref{fig:teaser}.

\xhdr{Notation}
Formally, we define a given input sensor stream as $x_i \in \mathbb{R}^{m \times 1}$, where $m$ is the length of the time series, and $x_{it}$ is the value of sensor stream $i$ at time $t$. We assemble $X=(x_0,..,x_n)\in \mathbb{R}^{m \times n}$ as a multivariate time series of $n$ streams in a given user's data.

\xhdr{Convolutional Encoder} The convolutional encoder learns a temporal, hierarchical feature representation of the raw sensor data.  Given the input multivariate time series $X$, we stack $q$ convolutional layers. In the simplest case, when stride and kernel size are not considered, the output of the $j^{th}$ channel $C_j$ with input size $\left(C_\mathrm{in},L_\mathrm{in}\right)$ is:

\begin{align*}
 \operatorname{out}\left(C_{j}\right)=\operatorname{bias}\left(C_{j}\right)+\sum_{k=0}^{C_{i n}-1} \operatorname{weight}\left(C_{j}, k\right) \star \operatorname{input}\left(k\right)
\end{align*}

where $\star$ is the cross-correlation operator, and $\operatorname{weight}$ and $\operatorname{bias}$ reflect learned parameters unique to each channel and each layer. Between layers we apply ReLU and batch-norm, which limit overfitting. We denote the final output of the CNN Encoder as $\overline{X}$, which has dimensionality $(C_{\mathrm{out},q},L_{\mathrm{out},q})$.

\xhdr{Transformer Blocks} Intuitively, this module learns relationships between the features produced in the final output of the CNN encoder. Our model uses a stack of $u$ transformer blocks, each composed of $r$ attention heads and a feed forward layer. We take the output of the final layer, $E$, to be the learned representation of the input time series:

\begin{align*}
h_0 &= \overline{X}^T + W_p \\
h_i &= \textrm{TransformerBlock}(h_{i-1}), i\in{1...u}\\ 
h^{norm}_i   &= \textrm{LayerNorm}(h_i)\\
E &= h^{norm}_{u}
\end{align*}

Where $W_p$ is a learned positional embedding matrix. 

\xhdr{Allowing for Missing Data} Researchers frequently report missingness as an obstacle to adopting deep learning techniques. In our dataset, 93\% of days contain missing data. Accordingly, we model missingness by replacing missing values with zeros and including a binary flag for each of the sensor streams which encodes if the sensor reading is missing in that timestep. 

\xhdr{Training} We train our model with the Adam optimizer \cite{kingma2017adam} and cross entropy loss. Details about hyperparameter tuning are available in Appendix~\ref{sec:hyperparam}.

\subsection{Self-Supervised Pretraining Tasks}
\label{subsec:self_supervised_tasks}
Much like in computer vision and NLP, labeled behavioral health data are expensive to collect at scale because labels require costly testing infrastructure. Transfer learning through self-supervised pretraining helps models learn generalizable representations from unlabeled data, where the status quo is only being able to learn from limited labled data. Here, we propose three techniques for self-supervised pretraining for behavioral data.

\xhdr{Same User} Prior work indicates that data from the same user on different days are often highly correlated relative to data from other users \cite{wang2016crosscheck}. Drawing inspiration from next sentence prediction tasks in NLP \cite{logeswaran_efficient_2018}, we hypothesize that a model that is trained to encode the differences between users may learn useful representations of behavioral data. To this end, we construct a dataset of one million pairs of (non-overlapping) windows from the same user, and one million pairs of windows from different users. We use the same encoder to generate embeddings for each of the windows in the pair, concatenate the embeddings, and use a linear layer to classify whether the pair of windows were from the same user (Figure~\ref{fig:teaser}A).

\xhdr{Autoencoder} For this pretraining task, we add a CNN decoder to the end of our model and use a mean-squared error objective to learn a reconstruction of the input time series from our model's lower dimensional embedding (Figure~\ref{fig:teaser}B). For simplicity's sake our decoder is a reflection of the encoder, i.e. it has the same architecture but with its one dimensional convolutions replaced with one dimensional deconvolutions and with a decreasing number of channels such that the final output has the same dimensionality as the original input. 

\xhdr{Domain Inspired Features} As previously mentioned, the majority of prior work in behavioral modeling has focused on classification tasks with hand-crafted features. While neural minute-level models may achieve superior performance than simple classifiers trained on these features, there is nonetheless a large body of work supporting the utility of handcrafted features in sensing~\cite{xu2021understanding,laport-lopez_review_2020,zhang_passive_2021, nair_understanding_2019, lin_healthwalks_2020, hafiz_wearable_2020, buda_outliers_2021, mairittha_crowdact_2021, meegahapola_one_2021}. For this pretraining task, we ask the model to perform a multiple regression to predict the daily features in Table~\ref{tab:features} on the final day of the seven day window (Figure~\ref{fig:teaser}C). Intuitively, there may be other, less obvious yet highly informative orthogonal features that our model could learn in order to reconstruct these higher level features. This task also has the added benefit of allowing us to inject expertise into the model. Since these features are calculated from the raw data (and in fact are mostly available through the FitBit API) and do not require any exogenous labels this task is fully self-supervised.

\section{Dataset}
\label{subsec:dataset}

Our dataset consists of 591k user-days of FitBit data collected from 5196 participants in the Homekit Flu Monitoring Study over the course of six months. Each minute the devices recorded the participant's total steps, average heart rate, and binary flags indicating if the participant was sleeping, awake, or in bed. Participants also completed daily surveys which asked if they were experiencing flu symptoms, including coughing, chills, fever, and fatigue. When a participant indicated that they were experiencing a cough and one other symptom, they were asked to self-administer a nasal swab test kit, which was then mailed to a lab for PCR analysis. Table~\ref{tab:data_stats} contains summary statistics for this study.
\section{Challenges in Evaluation}
\label{sec:evaluation}

There are few, if any, established best practices for evaluating behavioral models \cite{nestor2021dear,mcdermott2021reproducibility}. Given this lack of guidance, current evaluation paradigms vary significantly across studies. Here, we identify two common challenges to evaluating behavioral models in health and propose accompanying solutions, which we use to compare models in Section~\ref{sec:results}.
In summary, evaluations for behavioral models in healthcare should: 
\begin{itemize}
    \item Replicate genuine conditions, such as only using data from the past to inform predictions, and be tolerant to endemic missing data (Section~\ref{subsec:artificial_evaluation}).
    \item Faithfully quantify statistical significance, in particular when condition positive examples are rare, as is often the case in diagnostic testing (Section~\ref{subsec:problem_class_imbalance}). 
\end{itemize}

\subsection{Problem: Evaluations in artificial settings often lead to misleading performance estimates}
\label{subsec:artificial_evaluation}
Without a clear health application in mind from the outset it can be difficult for researchers to define tasks which faithfully replicate \qt{real world} conditions. It is not uncommon for models to:
\begin{itemize}
    \item train on data from the future \cite{wang2016crosscheck}, 
    \item use data collected in laboratory settings with limited ecological validity \cite{ismail_detection_2020}, 
    \item make predictions only if a user supplies sufficient data by using a device frequently \cite{malik_can_2020,wang2014studentlife}.
\end{itemize}
These practices may overestimate performance in diagnostic settings where a model would only have access to data from the past, rely on in-situ data, and would be most useful if it could function even with endemic missing data \cite{nestor2021dear, ismail_detection_2020}.  

\xhdr{Solution: Situate tasks around plausible healthcare scenarios} Here, we structure our prediction tasks to emulate the following realistic scenario:
\begin{quote}
Given training data from the first half of a flu season, how well can a model predict symptoms and infections in the second half of the flu season for every user on every day?
\end{quote}
Such a scenario arises in surveillance testing, where a population is frequently tested and positive individuals are asked to undertake additional testing or self isolate \cite{mercer_testing_2021}. Additionally, our tasks only use data from the seven days prior to a predicted event so that no information from the future informs a prediction about the past. We also include no explicit information about a users identity (e.g. participant id or demographics) to encourage models to learn generalizable motifs about activity data rather than facets of individual users' behavior. This evaluation setting follows existing best-practice recommendations and avoids falsely overstating the level of performance~\cite{nestor2021dear}.

\subsection{Problem: Predicting rare events limits statistical power and makes model selection inherently challenging} 
\label{subsec:problem_class_imbalance}
In mobile sensing for public health, relevant events are often fairly rare as intuitively most people are not sick most days. For example, one useful application of surveillance testing for respiratory viral infections is that if an individual tests positive they can self isolate and limit the spread of the infection to others. The CDC estimates that the average American has a 10\% chance of a symptomatic flu infection in a 365 day period, implying that the probability of an American receiving an initial positive flu diagnosis on a given day is roughly 0.027\% \cite{cdc_estimated_2021}. This corresponds to a 1:3,703 class imbalance, similar to the 1:2,760 in our evaluation dataset.

Modeling challenges aside, these extreme class imbalances make comparing model performance difficult as they limit statistical power and lead to large confidence intervals across many common test statistics. For example, for the Delong Test, a common test for comparing the ROC AUCs of two classfiers, the variance of the difference in AUCs is proportional to $\frac{1}{(N-m)m}$, where $N$ is the size of the dataset and $m$ is the number of true positive examples \cite{delong_comparing_1988}. Intuitively this variance is minimized, and statistical power maximized, when $m=N/2$ (a 1:1 class balance), and variance is maximized when $m=1$. 

Empirically, uncertainty can be quite high on realistic tasks, with peer studies of COVID-19 and flu detection reporting confidence intervals as high as $\pm0.1$ ROC AUC \cite{quer_wearable_2020}. Such statistical uncertainty makes it difficult to compare models, as extreme improvements in predictive performance on individual tasks are required to make strong claims about methodological progress.

As outlined in Section~\ref{subsec:artificial_evaluation}, many studies of interesting phenomena such as COVID-19 massively subsample true negatives to artificially deflate this class imbalance~\cite{quer_wearable_2020}. This creates a much simpler (but unrealistic) task, since higher false positive rates do not massively impact overall performance \cite{haibo_he_learning_2009}.

\xhdr{Solution: Aggregate performance across multiple tasks to increase statistical power} 
Here, rather than directly compare the performance of models on \emph{individual} tasks, we instead jointly compare the relative performance of models across \emph{all} tasks to improve statistical power. Intuitively, if a model performs best on all tasks, but not with high statistical significance on any one test, the probability that the model performance is indeed the same as all others is low. Specifically, we employ a Critical Difference plot \cite{carbonell_comparison_2000}, which first uses Friedman's statistic \cite{friedman_comparison_1940} to test the null hypothesis that there is no difference between the relative performance of models, and then deploys pairwise significance tests (e.g. Wilcoxon signed rank) between classifiers. This method, used here for the first time in mobile sensing for epidemiology, allows us to make statistically sound claims about our model's improvement over other common techniques without simplification of the underlying tasks (Section~\ref{sec:results}).  
\section{Empirical Evaluation}
\label{sec:results}
\mainTaskTable
\tableTaskFeatures

Here, we define five realistic prediction tasks and compare \modelname's performance against three representative baselines inspired by prior work (Section~\ref{subsec:pred_tasks}). 
In {\sc Experiment 1} we evaluate the performance between tasks through the framework defined in Section~\ref{sec:evaluation} to show that~\modelname~ outperforms these baselines. 
Next, {\sc Experiment 2} compares three novel pretraining methods for behavioral data to show that a method which integrates simple domain knowledge performs best (Section~\ref{subsec:results_pretraining}). 
{\sc Experiment 3 } in Section~\ref{subsec:sim_study} then shows that in simulated settings with limited training data, pretraining provides an average 0.04 ROC AUC performance boost relative to a non-pretrained model. Finally, we use a small, independently collected FitBit dataset to illustrate that features learned by our model on flu prediction generalize to COVID-19 prediction in a zero-shot task in {\sc Experiment 4}.

\subsection{{\sc Experiment 1:} Realistic Single Domain Prediction Tasks}
\label{subsec:pred_tasks}

We evaluate methods on five behavioral modeling tasks: 
\begin{itemize}
    \item \textbf{Flu Positivity:} Will the participant produce a nasal swab that tests positive for the flu today? This task emulates existing surveillance studies for both flu and COVID-19 where users are frequently tested for respiratory viral infection and asked to self-isolate in the event of a positive result \cite{chu_early_2020,fusco_covid-19_2020}.  
    \item \textbf{Severe Fever:} Will the participant report a severe fever (defined as three or more on a four point Likert scale) today?
    \item \textbf{Severe Cough:} Will the participant report a severe cough (defined as three or more on a four point Likert scale) today?
    \item \textbf{Severe Fatigue:} Will the participant report severe fatigue (defined as three or more on a four point Likert scale) today?
    \item \textbf{Flu Symptoms:} Will the participant report two or more flu symptoms (including cough, fever, and fatigue) of \emph{any} severity today? This prediction is important because preliminary screening for flu typically recommends a patient for additional treatment or testing if they report some combination of two or more symptoms \cite{cdc_influenza_symptoms_2021}, and this was the criterion used in the flu monitoring study that produced the evaluation dataset as well.
\end{itemize}

For these tasks, we follow our aforementioned evaluation best practices (Section~\ref{subsec:artificial_evaluation}) by training with data before the midpoint of the flu season (February 10th, in our case), and testing and evaluating models on data after the midpoint (as only data for one flu season is available). Furthermore, we make a prediction for every user on every day regardless of data quality, including predictions for users with no true positive labels.

In each case, we compare our model to the following baselines:
\begin{itemize}
    \item \textbf{XGBoost:} How well does our model perform relative to a non-neural baseline? Boosted decision trees are frequently used in many sensing studies because they are supported by common, easy to use libraries and often achieve strong performance out-of-the-box \cite{xu2021understanding}. Since boosted trees expectedly do not scale well to the thousands of observations in our raw time series data, we compute a set of commonly used features for each day in the window, and then concatenate these features for a final input. While neural models have surpassed non-neural classifiers in most CV and NLP applications, XGBoost is still commonly used in many contemporary sensing studies (e.g.,  \cite{zhang_passive_2021, nair_understanding_2019, lin_healthwalks_2020, hafiz_wearable_2020, buda_outliers_2021, mairittha_crowdact_2021, meegahapola_one_2021}). A list of all features is available in Table~\ref{tab:features}.
    
    \item \textbf{CNN:} How important are the transformer layers to our model's performance? To answer this question, we removed the transformer blocks from our model and passed the CNN's final output directly to a linear layer. 1D CNNs are frequently used in timeseries classification \cite{,pyrkov_extracting_2018,kiranyaz_1d_2021}, and have been applied to data from wearable devices before \cite{liu_fitbeat_2022,shen_ambulatory_2019, natarajan_assessment_2020}.
    
   \item \textbf{CNN-Transformer:} How important are pretraining and missingness flags to our model's performance? For this ablated model, we pass the CNN's final output to a transformer, but do not apply any pretraining method and do not include missingness flags. 
\end{itemize}

\noindent We do not include a \qt{transformer only} baseline (i.e., our model without the CNN encoder) because multi-head attention scales quadratically with the input length, making it computationally infeasible to perform such an experiment on a multi-day timeseries window (i.e., minute level data on a seven day window produces a 10,080 dimensional vector), which exceeds common context sizes in transformer models on commodity GPUs \cite{beltagy_longformer_2020}.

\xhdr{Results} Our model outperforms all baselines on every task (Table~\ref{tab:mainTasks}), which indicates that our method is a meaningful improvement over state of the art classifiers for behavioral data. Here we focus on precision-recall AUC, since the metric is typically more informative in cases of extreme class imbalance \cite{saito_precision-recall_2015}. Through Delong's test we find significant improvements in ROC AUC and PR AUC on the \qt{Severe Fatigue} and \qt{Flu Symptoms} tasks at $\alpha = 0.05$. As outlined in Section~\ref{sec:evaluation}, we employ Friedman's test and pair-wise Wilcoxon signed-rank tests to compare performance across tasks, and find that \modelname~significantly outranks XGBoost, CNNs, and CNN-Transformers at the best-practice parameter $\alpha=0.1$ \cite{carbonell_comparison_2000}, as it ranks first across all tasks. A critical difference plot is available in Figure~\ref{fig:CritDiff}, which shows that \modelname~is the best performing model overall, and that there is no statistically significant difference in the rankings of XGBoost, CNN, and the (non-pretrained) CNN-Transformer. We also experiment with the model's performance at modest levels of missing data, and find that it compares favorably to XGBoost (e.g. over 0.9 ROC AUC on the \qt{Flu Positivity} task even with 20\%-30\% of data missing; Figure~\ref{fig:binnedMissing}). A complete summary of results is available in Table~\ref{tab:mainTasks}. 

\figCritDiff

\subsection{{\sc Experiment 2:} Comparison of Self-Supervised Pretraining Methods}
\label{subsec:results_pretraining}
Next we compare the three pretraining techniques proposed in Section~\ref{subsec:self_supervised_tasks} on the \qt{Flu Symptoms} task (Section~\ref{subsec:pred_tasks}). This \qt{Flu Symptoms} task has the least extreme class imbalance (1:37) and therefore yields highest statistical power to differentiate model performance. We use the following pretraining method:

\begin{enumerate}
    \item Pretrain the model using all seven day windows in the train dataset. 
    \item Freeze the model's CNN and transformer layers. If the pretraining technique used a classification head, randomize its parameters. If instead a regression head was used, replace it with a randomly initialized classification head. 
    \item Finetune the model on the target task (\qt{Flu Symptoms}, in this case). 
\end{enumerate}

For all experiments, we use all of the model features described in Section~(\ref{subsec:model}) (i.e., the convolutional encoder, transformer blocks,  and missingness flags). For comparison, we include a \qt{No Pretraining} baseline, which shows the performance of a randomly initialized model. 

\xhdr{Results} ROC and Precision Recall curves for this experiment are available in Figure~\ref{fig:pretrainingResults}. \qt{Domain Inspired Features} pretraining, which trains the model to predict a pre-computed set of handcrafted features (Section~\ref{subsec:self_supervised_tasks}), significantly outperforms other pretraining techniques \emph{and} a randomly initialized model with a 16\% improvement in PR AUC. Notably, the model pretrained on the \qt{Same User} task does significantly worse than the others. 

We additionally compared this pretraining-fine tuning approach to a multitask learning where both the pretraining and the target prediction objective are optimized \emph{concurrently}. We repeated this comparison for each of the pretraining tasks~(Section~\ref{subsec:self_supervised_tasks}) in combination with the \qt{Flu Symptoms} task. This strategy produced no meaningful improvement over the randomly initialized model, indicating that unsupervised pretraining is a superior paradigm for this setting. 
In addition, the pretraining-finetuning paradigm enables us to separate these two steps across two datasets, especially when the target dataset is relatively small. This is the focus of the next two experiments.

\subsection{{\sc Experiment 3:} Transfer learning improves flu prediction performance on small datasets in a repeated simulation study}
\label{subsec:sim_study}
Labeled behavioral data is often prohibitively expensive to collect, particularly in the context of public health where ground-truth labels require costly testing infrastructure and study management. Accordingly, many studies from prior work operate on data with on the order of dozen participants \cite{xu2021understanding}. In this regard, one promising application of generalizable self-supervised pretraining is that models could leverage large unlabeled datasets to improve predictive power in settings with limited labeled training data. Here, we repeatedly simulate such settings to robustly investigate whether such transfer learning leads to performance improvements.

First, we isolate all 4,989 study participants who never tested positive for the flu. We treat this set as a large, unlabeled dataset which we use to pretrain our model on the self-supervised \qt{Daily Features} task (Section~\ref{subsec:self_supervised_tasks}). We then take the remaining 206 users who \emph{did} test positive at some point during the study, and randomly split this set into twenty folds of ten or eleven users each. This ensures that source and target domain share no participants in common. We provide an overview of this split in Figure~\ref{fig:pretrainedConcept}.

Next, for each fold we finetune the model on the supervised \qt{Flu Positivity} task using data from the fold, and evaluate it on the users in the remaining nineteen folds. We choose this task as it mirrors the zero shot setting in the external dataset of {\sc Experiment 4}. In both of these settings, all test subjects tested positive at some point and the predictive model attempts to predict on which day they do so.
This process simulates finetuning the model with fewer than a dozen users' data. We compare this approach to two non-pretrained models that only have access to the smaller target domain dataset: CNN-Transformer, and XGBoost trained on manually defined features (Table~\ref{tab:features}). 

\xhdr{Results} We find that our pretrained model outperforms non-pretrained models on the \qt{Flu Positivity} task when trained on fewer than a dozen participants (Figure~\ref{fig:flu_twenty_fold}). Pretraining alone increases average performance from  0.626 ROC AUC  to 0.665, and 0.017 PR AUC to 0.021 (both $p<0.05$, Mann-Whitney U). This indicates that Our Model can learn generalizable features from unlabeled training data.

\figPreTraining
\figFluTwentyFold
\label{subsec:results_transfer_learning}

\subsection{{\sc Experiment 4:} Pretrained behavioral representations enable zero-shot COVID-19 prediction in small external dataset}
\label{subsec:domain_transfer}
\domainTransferTable
Finally, we use a small, independently collected dataset of FitBit recordings and COVID-19 test results to show that \modelname~ can learn representations which generalize to entirely unseen diseases. This dataset contains 1470 total days of data for 32 individuals who tested positive with COVID-19 \cite{mishra_pre-symptomatic_2020}. The original study authors use a very different prediction task (retrospective prediction with no train/test split), and so it is not possible to make a direct comparison between our model and theirs, but this  dataset allows us to test our model's performance on an unseen disease. 

It is plausible that a self-supervised pretrained model, which in Experiment 1-3 showed good performance on flu related tasks, could support non-random predictive performance in a zero shot setting for COVID-19, as both are highly contagious respiratory viral infections which share symptoms and which may share a similar behavioral and physiological response to the disease (e.g., a change in resting heart rate around symptom onset~\cite{shapiro2021characterizing}). 

We pretrain our model with the \qt{Domain Inspired Features} task (Section~\ref{subsec:self_supervised_tasks}) and finetune it on the \qt{Predict Flu Positivity} task (Section~\ref{subsec:self_supervised_tasks}). Note that this is the same configuration as \qt{\modelname} in Table~\ref{tab:mainTasks}. Then, with no additional supervision we use the model to predict COVID-19 positivity for each user on each day in the small, external dataset. As a zero-shot baseline using the data, we calculate a set of day-level features (Table~\ref{tab:zero_shot_features}) from these data \qt{and} our original flu dataset (Section~\ref{subsec:dataset}) and train XGBoost using these features on the \qt{Flu Prediction} task. 
Note that neither Our Model nor the XGBoost baseline is exposed to \qt{any} data from the COVID-19 dataset during training. 

\xhdr{Results}
\modelname~outperforms XGBoost on this zero-shot task, achieving 0.68 ROC AUC, while XGBoost predicts at 0.51 ROC AUC (about random chance) (Table \ref{tab:domainTransfer}). This illustrates the feasibility of pretrained CNN-Transformers for novel disease prediction, and shows that the features learned by our model are general enough to effectively transfer between these domains. 

\section{Discussion \& Conclusion}
\label{sec:discussion}
In the COVID-19 pandemic rapid, cheap, over-the-counter antigen tests have been successfully deployed in surveillance testing, but the frequency of tests, more so than their sensitivity, remains a barrier to success in mitigating spread~\cite{larremore_test_2020}. In this paper, we outlined a framework for evaluating mobile sensing methods for respiratory viral infection prediction on realistic tasks (Section~{\ref{sec:evaluation}}) such as predicting flu positivity for every participant on every day of a flu season. While there are of course limitations to this work (e.g., a single flu season of data, limited sample size, restricted to flu and COVID-19 domains, limited interpretability of learned representations), we are able to report performance on par with COVID-19 rapid diagnostic tests, which have been found to have a ROC AUC of 0.84 (compared to 0.887 here on flu prediction; Table~\ref{tab:mainTasks}) in similar surveillance settings \cite{hirotsu_comparison_2020}. Certainly, more work is necessary to demonstrate similar performance levels in the same study population when deployed in a large-scale surveillance study. However, our findings demonstrate the potential of mobile sensing predictions as a complement to rapid antigen testing or as a trigger for additional testing. Specifically, our results indicate that pretraining, transformer self-attention, modeling missing data, and transfer learning are effective techniques in learning generalizable behavioral representations for mobile sensing. 
\balance

\bibliographystyle{ACM-Reference-Format} 
\bibliography{references_cleaned}


\begin{thebibliography}{70}


\ifx \showCODEN    \undefined \def \showCODEN     #1{\unskip}     \fi
\ifx \showDOI      \undefined \def \showDOI       #1{#1}\fi
\ifx \showISBNx    \undefined \def \showISBNx     #1{\unskip}     \fi
\ifx \showISBNxiii \undefined \def \showISBNxiii  #1{\unskip}     \fi
\ifx \showISSN     \undefined \def \showISSN      #1{\unskip}     \fi
\ifx \showLCCN     \undefined \def \showLCCN      #1{\unskip}     \fi
\ifx \shownote     \undefined \def \shownote      #1{#1}          \fi
\ifx \showarticletitle \undefined \def \showarticletitle #1{#1}   \fi
\ifx \showURL      \undefined \def \showURL       {\relax}        \fi
\providecommand\bibfield[2]{#2}
\providecommand\bibinfo[2]{#2}
\providecommand\natexlab[1]{#1}
\providecommand\showeprint[2][]{arXiv:#2}

\bibitem[\protect\citeauthoryear{??}{cdc}{2021a}]%
        {cdc_estimated_2021}
 \bibinfo{year}{2021}\natexlab{a}.
\newblock \bibinfo{title}{Estimated {Flu}-{Related} {Illnesses}, {Medical}
  visits, {Hospitalizations}, and {Deaths} in the {United} {States} —
  2018–2019 {Flu} {Season} {\textbar} {CDC}}.
\newblock
\newblock


\bibitem[\protect\citeauthoryear{??}{cdc}{2021b}]%
        {cdc_influenza_symptoms_2021}
 \bibinfo{year}{2021}\natexlab{b}.
\newblock \bibinfo{title}{Influenza {Signs} and {Symptoms} and the {Role} of
  {Laboratory} {Diagnostics}}.
\newblock
\newblock


\bibitem[\protect\citeauthoryear{Bachman, Grant, Anderson, Alonzo, Garing,
  et~al\mbox{.}}{Bachman et~al\mbox{.}}{2021}]%
        {bachman_clinical_2021}
\bibfield{author}{\bibinfo{person}{Christine~M. Bachman},
  \bibinfo{person}{Benjamin~D. Grant}, \bibinfo{person}{Caitlin~E. Anderson},
  \bibinfo{person}{Luis~F. Alonzo}, \bibinfo{person}{Spencer Garing},
  {et~al\mbox{.}}} \bibinfo{year}{2021}\natexlab{}.
\newblock \showarticletitle{Clinical validation of an open-access
  {SARS}-{COV}-2 antigen detection lateral flow assay, compared to commercially
  available assays}.
\newblock \bibinfo{journal}{\emph{PLOS ONE}} (\bibinfo{year}{2021}).
\newblock
\showISSN{1932-6203}


\bibitem[\protect\citeauthoryear{Beltagy, Peters, and Cohan}{Beltagy
  et~al\mbox{.}}{2020}]%
        {beltagy_longformer_2020}
\bibfield{author}{\bibinfo{person}{Iz Beltagy}, \bibinfo{person}{Matthew~E.
  Peters}, {and} \bibinfo{person}{Arman Cohan}.}
  \bibinfo{year}{2020}\natexlab{}.
\newblock \showarticletitle{Longformer: {The} {Long}-{Document} {Transformer}}.
\newblock \bibinfo{journal}{\emph{arXiv:2004.05150 [cs]}}
  (\bibinfo{year}{2020}).
\newblock


\bibitem[\protect\citeauthoryear{Brazdil and Soares}{Brazdil and
  Soares}{2000}]%
        {carbonell_comparison_2000}
\bibfield{author}{\bibinfo{person}{Pavel~B. Brazdil} {and}
  \bibinfo{person}{Carlos Soares}.} \bibinfo{year}{2000}\natexlab{}.
\newblock \showarticletitle{A {Comparison} of {Ranking} {Methods} for
  {Classification} {Algorithm} {Selection}}.
\newblock In \bibinfo{booktitle}{\emph{Machine {Learning}: {ECML} 2000}},
  \bibfield{editor}{\bibinfo{person}{Jaime~G. Carbonell},
  \bibinfo{person}{Jörg Siekmann}, \bibinfo{person}{G.~Goos},
  \bibinfo{person}{J.~Hartmanis}, \bibinfo{person}{J.~van Leeuwen},
  {et~al\mbox{.}}} (Eds.). \bibinfo{publisher}{Springer Berlin Heidelberg},
  \bibinfo{address}{Berlin, Heidelberg}.
\newblock


\bibitem[\protect\citeauthoryear{Brook, Northrup, Ehrenberg, Doudna, and
  Boots}{Brook et~al\mbox{.}}{2021}]%
        {brook_optimizing_2021}
\bibfield{author}{\bibinfo{person}{Cara~E. Brook}, \bibinfo{person}{Graham~R.
  Northrup}, \bibinfo{person}{Alexander~J. Ehrenberg},
  \bibinfo{person}{Jennifer~A. Doudna}, {and} \bibinfo{person}{Mike Boots}.}
  \bibinfo{year}{2021}\natexlab{}.
\newblock \showarticletitle{Optimizing {COVID}-19 control with asymptomatic
  surveillance testing in a university environment}.
\newblock \bibinfo{journal}{\emph{Epidemics}} (\bibinfo{year}{2021}).
\newblock


\bibitem[\protect\citeauthoryear{Buda, Khwaja, and Matic}{Buda
  et~al\mbox{.}}{2021}]%
        {buda_outliers_2021}
\bibfield{author}{\bibinfo{person}{Teodora~Sandra Buda},
  \bibinfo{person}{Mohammed Khwaja}, {and} \bibinfo{person}{Aleksandar Matic}.}
  \bibinfo{year}{2021}\natexlab{}.
\newblock \showarticletitle{Outliers in {Smartphone} {Sensor} {Data} {Reveal}
  {Outliers} in {Daily} {Happiness}}.
\newblock \bibinfo{journal}{\emph{Proceedings of the ACM on Interactive,
  Mobile, Wearable and Ubiquitous Technologies}} (\bibinfo{year}{2021}).
\newblock
\showISSN{2474-9567}


\bibitem[\protect\citeauthoryear{Cao, Zheng, Zhang, Yu, Piscitello,
  et~al\mbox{.}}{Cao et~al\mbox{.}}{2017}]%
        {cao2017deepmood}
\bibfield{author}{\bibinfo{person}{Bokai Cao}, \bibinfo{person}{Lei Zheng},
  \bibinfo{person}{Chenwei Zhang}, \bibinfo{person}{Philip~S Yu},
  \bibinfo{person}{Andrea Piscitello}, {et~al\mbox{.}}}
  \bibinfo{year}{2017}\natexlab{}.
\newblock \showarticletitle{Deepmood: modeling mobile phone typing dynamics for
  mood detection}. In \bibinfo{booktitle}{\emph{KDD}}.
\newblock


\bibitem[\protect\citeauthoryear{Chen, Lundberg, Erion, Kim, and Lee}{Chen
  et~al\mbox{.}}{2020}]%
        {chen2020forecasting}
\bibfield{author}{\bibinfo{person}{Hugh Chen}, \bibinfo{person}{Scott
  Lundberg}, \bibinfo{person}{Gabe Erion}, \bibinfo{person}{Jerry~H Kim}, {and}
  \bibinfo{person}{Su-In Lee}.} \bibinfo{year}{2020}\natexlab{}.
\newblock \showarticletitle{Forecasting adverse surgical events using
  self-supervised transfer learning for physiological signals}.
\newblock \bibinfo{journal}{\emph{arXiv:2002.04770}} (\bibinfo{year}{2020}).
\newblock


\bibitem[\protect\citeauthoryear{Chu, Lofgren, Halloran, Kuan, Hudgens,
  et~al\mbox{.}}{Chu et~al\mbox{.}}{2012}]%
        {chu_performance_2012}
\bibfield{author}{\bibinfo{person}{Haitao Chu}, \bibinfo{person}{Eric~T.
  Lofgren}, \bibinfo{person}{M.~Elizabeth Halloran}, \bibinfo{person}{Pei~F.
  Kuan}, \bibinfo{person}{Michael Hudgens}, {et~al\mbox{.}}}
  \bibinfo{year}{2012}\natexlab{}.
\newblock \showarticletitle{Performance of rapid influenza {H1N1} diagnostic
  tests: a meta-analysis}.
\newblock \bibinfo{journal}{\emph{Influenza and Other Respiratory Viruses}}
  (\bibinfo{year}{2012}).
\newblock


\bibitem[\protect\citeauthoryear{Chu, Englund, Starita, Famulare, Brandstetter,
  et~al\mbox{.}}{Chu et~al\mbox{.}}{2020}]%
        {chu_early_2020}
\bibfield{author}{\bibinfo{person}{Helen~Y. Chu}, \bibinfo{person}{Janet~A.
  Englund}, \bibinfo{person}{Lea~M. Starita}, \bibinfo{person}{Michael
  Famulare}, \bibinfo{person}{Elisabeth Brandstetter}, {et~al\mbox{.}}}
  \bibinfo{year}{2020}\natexlab{}.
\newblock \showarticletitle{Early {Detection} of {Covid}-19 through a
  {Citywide} {Pandemic} {Surveillance} {Platform}}.
\newblock \bibinfo{journal}{\emph{New England Journal of Medicine}}
  (\bibinfo{year}{2020}).
\newblock


\bibitem[\protect\citeauthoryear{DeLong, DeLong, and Clarke-Pearson}{DeLong
  et~al\mbox{.}}{1988}]%
        {delong_comparing_1988}
\bibfield{author}{\bibinfo{person}{Elizabeth~R. DeLong},
  \bibinfo{person}{David~M. DeLong}, {and} \bibinfo{person}{Daniel~L.
  Clarke-Pearson}.} \bibinfo{year}{1988}\natexlab{}.
\newblock \showarticletitle{Comparing the {Areas} under {Two} or {More}
  {Correlated} {Receiver} {Operating} {Characteristic} {Curves}: {A}
  {Nonparametric} {Approach}}.
\newblock \bibinfo{journal}{\emph{Biometrics}} (\bibinfo{year}{1988}).
\newblock


\bibitem[\protect\citeauthoryear{Devlin, Chang, Lee, and Toutanova}{Devlin
  et~al\mbox{.}}{2019}]%
        {devlin2019bert}
\bibfield{author}{\bibinfo{person}{Jacob Devlin}, \bibinfo{person}{Ming-Wei
  Chang}, \bibinfo{person}{Kenton Lee}, {and} \bibinfo{person}{Kristina
  Toutanova}.} \bibinfo{year}{2019}\natexlab{}.
\newblock \showarticletitle{BERT: Pre-training of Deep Bidirectional
  Transformers for Language Understanding}. In
  \bibinfo{booktitle}{\emph{NAACL-HLT (1)}}.
\newblock


\bibitem[\protect\citeauthoryear{Fawaz, Forestier, Weber, Idoumghar, and
  Muller}{Fawaz et~al\mbox{.}}{2019}]%
        {fawaz_deep_2019}
\bibfield{author}{\bibinfo{person}{Hassan~Ismail Fawaz},
  \bibinfo{person}{Germain Forestier}, \bibinfo{person}{Jonathan Weber},
  \bibinfo{person}{Lhassane Idoumghar}, {and} \bibinfo{person}{Pierre-Alain
  Muller}.} \bibinfo{year}{2019}\natexlab{}.
\newblock \showarticletitle{Deep learning for time series classification: a
  review}.
\newblock  \bibinfo{volume}{33}, \bibinfo{number}{4} (\bibinfo{date}{July}
  \bibinfo{year}{2019}), \bibinfo{pages}{917--963}.
\newblock
\showISSN{1384-5810, 1573-756X}
\newblock
\shownote{arXiv: 1809.04356.}


\bibitem[\protect\citeauthoryear{Friedman}{Friedman}{1940}]%
        {friedman_comparison_1940}
\bibfield{author}{\bibinfo{person}{Milton Friedman}.}
  \bibinfo{year}{1940}\natexlab{}.
\newblock \showarticletitle{A {Comparison} of {Alternative} {Tests} of
  {Significance} for the {Problem} of \$m\$ {Rankings}}.
\newblock  (\bibinfo{year}{1940}), \bibinfo{pages}{86--92}.
\newblock
\showISSN{0003-4851, 2168-8990}


\bibitem[\protect\citeauthoryear{Fusco, Pisaturo, Iodice, Bellopede, Tambaro,
  et~al\mbox{.}}{Fusco et~al\mbox{.}}{2020}]%
        {fusco_covid-19_2020}
\bibfield{author}{\bibinfo{person}{F.~M. Fusco}, \bibinfo{person}{M. Pisaturo},
  \bibinfo{person}{V. Iodice}, \bibinfo{person}{R. Bellopede},
  \bibinfo{person}{O. Tambaro}, {et~al\mbox{.}}}
  \bibinfo{year}{2020}\natexlab{}.
\newblock \showarticletitle{{COVID}-19 among healthcare workers in a specialist
  infectious diseases setting in {Naples}, {Southern} {Italy}: results of a
  cross-sectional surveillance study}.
\newblock \bibinfo{journal}{\emph{Journal of Hospital Infection}}
  (\bibinfo{year}{2020}).
\newblock
\showISSN{0195-6701}


\bibitem[\protect\citeauthoryear{Hafiz, Miskowiak, Maxhuni, Kessing, and
  Bardram}{Hafiz et~al\mbox{.}}{2020}]%
        {hafiz_wearable_2020}
\bibfield{author}{\bibinfo{person}{Pegah Hafiz},
  \bibinfo{person}{Kamilla~Woznica Miskowiak}, \bibinfo{person}{Alban Maxhuni},
  \bibinfo{person}{Lars~Vedel Kessing}, {and} \bibinfo{person}{Jakob~Eyvind
  Bardram}.} \bibinfo{year}{2020}\natexlab{}.
\newblock \showarticletitle{Wearable {Computing} {Technology} for {Assessment}
  of {Cognitive} {Functioning} of {Bipolar} {Patients} and {Healthy}
  {Controls}}.
\newblock \bibinfo{journal}{\emph{IMWUT}} (\bibinfo{year}{2020}).
\newblock


\bibitem[\protect\citeauthoryear{{Haibo He} and Garcia}{{Haibo He} and
  Garcia}{2009}]%
        {haibo_he_learning_2009}
\bibfield{author}{\bibinfo{person}{{Haibo He}} {and} \bibinfo{person}{E.A.
  Garcia}.} \bibinfo{year}{2009}\natexlab{}.
\newblock \showarticletitle{Learning from {Imbalanced} {Data}}.
\newblock \bibinfo{journal}{\emph{IEEE Transactions on Knowledge and Data
  Engineering}} (\bibinfo{year}{2009}), \bibinfo{pages}{1263--1284}.
\newblock
\showISSN{1041-4347}


\bibitem[\protect\citeauthoryear{Hallgr{\'\i}msson, Jankovic, Althoff, and
  Foschini}{Hallgr{\'\i}msson et~al\mbox{.}}{2018}]%
        {hallgrimsson2018learning}
\bibfield{author}{\bibinfo{person}{Haraldur~T Hallgr{\'\i}msson},
  \bibinfo{person}{Filip Jankovic}, \bibinfo{person}{Tim Althoff}, {and}
  \bibinfo{person}{Luca Foschini}.} \bibinfo{year}{2018}\natexlab{}.
\newblock \showarticletitle{Learning individualized cardiovascular responses
  from large-scale wearable sensors data}.
\newblock \bibinfo{journal}{\emph{NeurIPS ML4H}} (\bibinfo{year}{2018}).
\newblock


\bibitem[\protect\citeauthoryear{He, Zhang, Ren, and Sun}{He
  et~al\mbox{.}}{2016}]%
        {he2016resnet}
\bibfield{author}{\bibinfo{person}{Kaiming He}, \bibinfo{person}{Xiangyu
  Zhang}, \bibinfo{person}{Shaoqing Ren}, {and} \bibinfo{person}{Jian Sun}.}
  \bibinfo{year}{2016}\natexlab{}.
\newblock \showarticletitle{Deep residual learning for image recognition}. In
  \bibinfo{booktitle}{\emph{Proceedings of the IEEE conference on computer
  vision and pattern recognition}}. \bibinfo{pages}{770--778}.
\newblock


\bibitem[\protect\citeauthoryear{Hirotsu, Maejima, Shibusawa, Nagakubo, Hosaka,
  et~al\mbox{.}}{Hirotsu et~al\mbox{.}}{2020}]%
        {hirotsu_comparison_2020}
\bibfield{author}{\bibinfo{person}{Yosuke Hirotsu}, \bibinfo{person}{Makoto
  Maejima}, \bibinfo{person}{Masahiro Shibusawa}, \bibinfo{person}{Yuki
  Nagakubo}, \bibinfo{person}{Kazuhiro Hosaka}, {et~al\mbox{.}}}
  \bibinfo{year}{2020}\natexlab{}.
\newblock \showarticletitle{Comparison of automated {SARS}-{CoV}-2 antigen test
  for {COVID}-19 infection with quantitative {RT}-{PCR} using 313
  nasopharyngeal swabs, including from seven serially followed patients}.
\newblock \bibinfo{journal}{\emph{International Journal of Infectious
  Diseases}}  \bibinfo{volume}{99} (\bibinfo{date}{Oct.} \bibinfo{year}{2020}),
  \bibinfo{pages}{397--402}.
\newblock
\showISSN{1201-9712}


\bibitem[\protect\citeauthoryear{Hochreiter and Schmidhuber}{Hochreiter and
  Schmidhuber}{1997}]%
        {hochreiter1997long}
\bibfield{author}{\bibinfo{person}{Sepp Hochreiter} {and}
  \bibinfo{person}{J{\"u}rgen Schmidhuber}.} \bibinfo{year}{1997}\natexlab{}.
\newblock \showarticletitle{Long short-term memory}.
\newblock \bibinfo{journal}{\emph{Neural computation}} (\bibinfo{year}{1997}),
  \bibinfo{pages}{1735--1780}.
\newblock


\bibitem[\protect\citeauthoryear{{International Monetary Fund}}{{International
  Monetary Fund}}{2020}]%
        {imf2020covid}
\bibfield{author}{\bibinfo{person}{{International Monetary Fund}}.}
  \bibinfo{year}{2020}\natexlab{}.
\newblock \bibinfo{title}{{A Crisis Like No Other, An Uncertain Recovery}}.
\newblock
\newblock


\bibitem[\protect\citeauthoryear{Ismail, Deshmukh, and Singh}{Ismail
  et~al\mbox{.}}{2020}]%
        {ismail_detection_2020}
\bibfield{author}{\bibinfo{person}{Mahmoud~Al Ismail}, \bibinfo{person}{Soham
  Deshmukh}, {and} \bibinfo{person}{Rita Singh}.}
  \bibinfo{year}{2020}\natexlab{}.
\newblock \showarticletitle{Detection of {COVID}-19 through the analysis of
  vocal fold oscillations}.
\newblock  (\bibinfo{year}{2020}).
\newblock


\bibitem[\protect\citeauthoryear{Jaques, Rudovic, Taylor, Sano, and
  Picard}{Jaques et~al\mbox{.}}{2017}]%
        {jaques2017multitask}
\bibfield{author}{\bibinfo{person}{Natasha Jaques},
  \bibinfo{person}{Ognjen~(Oggi) Rudovic}, \bibinfo{person}{Sara Taylor},
  \bibinfo{person}{Akane Sano}, {and} \bibinfo{person}{Rosalind Picard}.}
  \bibinfo{year}{2017}\natexlab{}.
\newblock \showarticletitle{Predicting Tomorrow’s Mood, Health, and Stress
  Level using Personalized Multitask Learning and Domain Adaptation}. In
  \bibinfo{booktitle}{\emph{Proceedings of IJCAI 2017 Workshop on Artificial
  Intelligence in Affective Computing}}.
\newblock


\bibitem[\protect\citeauthoryear{Kingma and Ba}{Kingma and Ba}{2017}]%
        {kingma2017adam}
\bibfield{author}{\bibinfo{person}{Diederik~P. Kingma} {and}
  \bibinfo{person}{Jimmy Ba}.} \bibinfo{year}{2017}\natexlab{}.
\newblock \bibinfo{title}{Adam: A Method for Stochastic Optimization}.
\newblock
\newblock


\bibitem[\protect\citeauthoryear{Kiranyaz, Avci, Abdeljaber, Ince, Gabbouj,
  et~al\mbox{.}}{Kiranyaz et~al\mbox{.}}{2021}]%
        {kiranyaz_1d_2021}
\bibfield{author}{\bibinfo{person}{Serkan Kiranyaz}, \bibinfo{person}{Onur
  Avci}, \bibinfo{person}{Osama Abdeljaber}, \bibinfo{person}{Turker Ince},
  \bibinfo{person}{Moncef Gabbouj}, {et~al\mbox{.}}}
  \bibinfo{year}{2021}\natexlab{}.
\newblock \showarticletitle{{1D} convolutional neural networks and
  applications: {A} survey}.
\newblock \bibinfo{journal}{\emph{Mechanical Systems and Signal Processing}}
  (\bibinfo{year}{2021}).
\newblock


\bibitem[\protect\citeauthoryear{Kolbeinsson, Gade, Kainkaryam, Jankovic, and
  Foschini}{Kolbeinsson et~al\mbox{.}}{2021}]%
        {kolbeinsson_self-supervision_2021}
\bibfield{author}{\bibinfo{person}{Arinbjörn Kolbeinsson},
  \bibinfo{person}{Piyusha Gade}, \bibinfo{person}{Raghu Kainkaryam},
  \bibinfo{person}{Filip Jankovic}, {and} \bibinfo{person}{Luca Foschini}.}
  \bibinfo{year}{2021}\natexlab{}.
\newblock \showarticletitle{Self-supervision of wearable sensors time-series
  data for influenza detection}.
\newblock \bibinfo{journal}{\emph{arXiv:2112.13755 [cs]}} (\bibinfo{date}{Dec.}
  \bibinfo{year}{2021}).
\newblock


\bibitem[\protect\citeauthoryear{Krizhevsky, Sutskever, and Hinton}{Krizhevsky
  et~al\mbox{.}}{2012}]%
        {krizhevsky2012alexnet}
\bibfield{author}{\bibinfo{person}{Alex Krizhevsky}, \bibinfo{person}{Ilya
  Sutskever}, {and} \bibinfo{person}{Geoffrey~E Hinton}.}
  \bibinfo{year}{2012}\natexlab{}.
\newblock \showarticletitle{Imagenet classification with deep convolutional
  neural networks}.
\newblock \bibinfo{journal}{\emph{NeurIPS}} (\bibinfo{year}{2012}).
\newblock


\bibitem[\protect\citeauthoryear{Lan, Chen, Goodman, Gimpel, Sharma,
  et~al\mbox{.}}{Lan et~al\mbox{.}}{2019}]%
        {lan2019albert}
\bibfield{author}{\bibinfo{person}{Zhenzhong Lan}, \bibinfo{person}{Mingda
  Chen}, \bibinfo{person}{Sebastian Goodman}, \bibinfo{person}{Kevin Gimpel},
  \bibinfo{person}{Piyush Sharma}, {et~al\mbox{.}}}
  \bibinfo{year}{2019}\natexlab{}.
\newblock \showarticletitle{Albert: A lite bert for self-supervised learning of
  language representations}.
\newblock \bibinfo{journal}{\emph{arXiv preprint arXiv:1909.11942}}
  (\bibinfo{year}{2019}).
\newblock


\bibitem[\protect\citeauthoryear{Laport-López, Serrano, Bajo, and
  Campbell}{Laport-López et~al\mbox{.}}{2020}]%
        {laport-lopez_review_2020}
\bibfield{author}{\bibinfo{person}{Francisco Laport-López},
  \bibinfo{person}{Emilio Serrano}, \bibinfo{person}{Javier Bajo}, {and}
  \bibinfo{person}{Andrew~T. Campbell}.} \bibinfo{year}{2020}\natexlab{}.
\newblock \showarticletitle{A review of mobile sensing systems, applications,
  and opportunities}.
\newblock \bibinfo{journal}{\emph{Knowledge and Information Systems}}
  (\bibinfo{year}{2020}).
\newblock


\bibitem[\protect\citeauthoryear{Larremore, Wilder, Lester, Shehata, Burke,
  et~al\mbox{.}}{Larremore et~al\mbox{.}}{2020}]%
        {larremore_test_2020}
\bibfield{author}{\bibinfo{person}{Daniel~B. Larremore}, \bibinfo{person}{Bryan
  Wilder}, \bibinfo{person}{Evan Lester}, \bibinfo{person}{Soraya Shehata},
  \bibinfo{person}{James~M. Burke}, {et~al\mbox{.}}}
  \bibinfo{year}{2020}\natexlab{}.
\newblock \showarticletitle{Test sensitivity is secondary to frequency and
  turnaround time for {COVID}-19 surveillance}.
\newblock \bibinfo{journal}{\emph{medRxiv}} (\bibinfo{date}{Sept.}
  \bibinfo{year}{2020}).
\newblock


\bibitem[\protect\citeauthoryear{Lewis, Liu, Goyal, Ghazvininejad, Mohamed,
  et~al\mbox{.}}{Lewis et~al\mbox{.}}{2019}]%
        {lewis2019bart}
\bibfield{author}{\bibinfo{person}{Mike Lewis}, \bibinfo{person}{Yinhan Liu},
  \bibinfo{person}{Naman Goyal}, \bibinfo{person}{Marjan Ghazvininejad},
  \bibinfo{person}{Abdelrahman Mohamed}, {et~al\mbox{.}}}
  \bibinfo{year}{2019}\natexlab{}.
\newblock \showarticletitle{Bart: Denoising sequence-to-sequence pre-training
  for natural language generation, translation, and comprehension}.
\newblock  (\bibinfo{year}{2019}).
\newblock


\bibitem[\protect\citeauthoryear{Li and Sano}{Li and Sano}{2020}]%
        {li2020extraction}
\bibfield{author}{\bibinfo{person}{Boning Li} {and} \bibinfo{person}{Akane
  Sano}.} \bibinfo{year}{2020}\natexlab{}.
\newblock \showarticletitle{Extraction and Interpretation of Deep
  Autoencoder-based Temporal Features from Wearables for Forecasting
  Personalized Mood, Health, and Stress}.
\newblock \bibinfo{journal}{\emph{IMWUT}} (\bibinfo{year}{2020}).
\newblock


\bibitem[\protect\citeauthoryear{Lin, Lyu, Cao, Xu, Wei, et~al\mbox{.}}{Lin
  et~al\mbox{.}}{2020}]%
        {lin_healthwalks_2020}
\bibfield{author}{\bibinfo{person}{Zongyu Lin}, \bibinfo{person}{Shiqing Lyu},
  \bibinfo{person}{Hancheng Cao}, \bibinfo{person}{Fengli Xu},
  \bibinfo{person}{Yuqiong Wei}, {et~al\mbox{.}}}
  \bibinfo{year}{2020}\natexlab{}.
\newblock \showarticletitle{{HealthWalks}: {Sensing} {Fine}-grained
  {Individual} {Health} {Condition} via {Mobility} {Data}}.
\newblock \bibinfo{journal}{\emph{IMWUT}} (\bibinfo{year}{2020}),
  \bibinfo{pages}{1--26}.
\newblock


\bibitem[\protect\citeauthoryear{Liu, Han, Puyal, Kontaxis, Sun,
  et~al\mbox{.}}{Liu et~al\mbox{.}}{2022}]%
        {liu_fitbeat_2022}
\bibfield{author}{\bibinfo{person}{Shuo Liu}, \bibinfo{person}{Jing Han},
  \bibinfo{person}{Estela~Laporta Puyal}, \bibinfo{person}{Spyridon Kontaxis},
  \bibinfo{person}{Shaoxiong Sun}, {et~al\mbox{.}}}
  \bibinfo{year}{2022}\natexlab{}.
\newblock \showarticletitle{Fitbeat: {COVID}-19 estimation based on wristband
  heart rate using a contrastive convolutional auto-encoder}.
\newblock \bibinfo{journal}{\emph{Pattern Recognition}} (\bibinfo{year}{2022}).
\newblock


\bibitem[\protect\citeauthoryear{Liu, Ott, Goyal, Du, Joshi, et~al\mbox{.}}{Liu
  et~al\mbox{.}}{2019}]%
        {liu2019roberta}
\bibfield{author}{\bibinfo{person}{Yinhan Liu}, \bibinfo{person}{Myle Ott},
  \bibinfo{person}{Naman Goyal}, \bibinfo{person}{Jingfei Du},
  \bibinfo{person}{Mandar Joshi}, {et~al\mbox{.}}}
  \bibinfo{year}{2019}\natexlab{}.
\newblock \showarticletitle{Roberta: A robustly optimized bert pretraining
  approach}.
\newblock \bibinfo{journal}{\emph{arXiv:1907.11692}} (\bibinfo{year}{2019}).
\newblock


\bibitem[\protect\citeauthoryear{Logeswaran and Lee}{Logeswaran and
  Lee}{2018}]%
        {logeswaran_efficient_2018}
\bibfield{author}{\bibinfo{person}{Lajanugen Logeswaran} {and}
  \bibinfo{person}{Honglak Lee}.} \bibinfo{year}{2018}\natexlab{}.
\newblock \showarticletitle{An efficient framework for learning sentence
  representations}.
\newblock


\bibitem[\protect\citeauthoryear{Ma, Campbell, Cook, Lach, Patel,
  et~al\mbox{.}}{Ma et~al\mbox{.}}{2020}]%
        {ma2020transfer}
\bibfield{author}{\bibinfo{person}{Yuchao Ma}, \bibinfo{person}{Andrew~T
  Campbell}, \bibinfo{person}{Diane~J Cook}, \bibinfo{person}{John Lach},
  \bibinfo{person}{Shwetak~N Patel}, {et~al\mbox{.}}}
  \bibinfo{year}{2020}\natexlab{}.
\newblock \showarticletitle{Transfer learning for activity recognition in
  mobile health}.
\newblock  (\bibinfo{year}{2020}).
\newblock


\bibitem[\protect\citeauthoryear{Mairittha, Mairittha, Lago, and
  Inoue}{Mairittha et~al\mbox{.}}{2021}]%
        {mairittha_crowdact_2021}
\bibfield{author}{\bibinfo{person}{Nattaya Mairittha}, \bibinfo{person}{Tittaya
  Mairittha}, \bibinfo{person}{Paula Lago}, {and} \bibinfo{person}{Sozo
  Inoue}.} \bibinfo{year}{2021}\natexlab{}.
\newblock \showarticletitle{{CrowdAct}: {Achieving} {High}-{Quality}
  {Crowdsourced} {Datasets} in {Mobile} {Activity} {Recognition}}.
\newblock \bibinfo{journal}{\emph{IMWUT}} (\bibinfo{year}{2021}).
\newblock


\bibitem[\protect\citeauthoryear{Malik, Doryab, Merrill, Pfeffer, and
  Dey}{Malik et~al\mbox{.}}{2020}]%
        {malik_can_2020}
\bibfield{author}{\bibinfo{person}{Momin~M. Malik}, \bibinfo{person}{Afsaneh
  Doryab}, \bibinfo{person}{Michael Merrill}, \bibinfo{person}{Jürgen
  Pfeffer}, {and} \bibinfo{person}{Anind~K. Dey}.}
  \bibinfo{year}{2020}\natexlab{}.
\newblock \showarticletitle{Can {Smartphone} {Co}-locations {Detect}
  {Friendship}? {It} {Depends} {How} {You} {Model} {It}}.
\newblock \bibinfo{journal}{\emph{arXiv:2008.02919 [cs]}}
  (\bibinfo{year}{2020}).
\newblock


\bibitem[\protect\citeauthoryear{McDermott, Wang, Marinsek, Ranganath,
  Foschini, et~al\mbox{.}}{McDermott et~al\mbox{.}}{2021}]%
        {mcdermott2021reproducibility}
\bibfield{author}{\bibinfo{person}{Matthew~BA McDermott},
  \bibinfo{person}{Shirly Wang}, \bibinfo{person}{Nikki Marinsek},
  \bibinfo{person}{Rajesh Ranganath}, \bibinfo{person}{Luca Foschini},
  {et~al\mbox{.}}} \bibinfo{year}{2021}\natexlab{}.
\newblock \showarticletitle{Reproducibility in machine learning for health
  research: Still a ways to go}.
\newblock \bibinfo{journal}{\emph{Science Translational Medicine}}
  (\bibinfo{year}{2021}).
\newblock


\bibitem[\protect\citeauthoryear{Meegahapola, Ruiz-Correa, Robledo-Valero,
  Hernandez-Huerfano, Alvarez-Rivera, et~al\mbox{.}}{Meegahapola
  et~al\mbox{.}}{2021}]%
        {meegahapola_one_2021}
\bibfield{author}{\bibinfo{person}{Lakmal Meegahapola},
  \bibinfo{person}{Salvador Ruiz-Correa}, \bibinfo{person}{Viridiana del~Carmen
  Robledo-Valero}, \bibinfo{person}{Emilio~Ernesto Hernandez-Huerfano},
  \bibinfo{person}{Leonardo Alvarez-Rivera}, {et~al\mbox{.}}}
  \bibinfo{year}{2021}\natexlab{}.
\newblock \showarticletitle{One {More} {Bite}? {Inferring} {Food} {Consumption}
  {Level} of {College} {Students} {Using} {Smartphone} {Sensing} and
  {Self}-{Reports}}.
\newblock \bibinfo{journal}{\emph{IMWUT}} (\bibinfo{year}{2021}).
\newblock


\bibitem[\protect\citeauthoryear{Mercer and Salit}{Mercer and Salit}{2021}]%
        {mercer_testing_2021}
\bibfield{author}{\bibinfo{person}{Tim~R. Mercer} {and} \bibinfo{person}{Marc
  Salit}.} \bibinfo{year}{2021}\natexlab{}.
\newblock \showarticletitle{Testing at scale during the {COVID}-19 pandemic}.
\newblock \bibinfo{journal}{\emph{Nature Reviews Genetics}}
  (\bibinfo{year}{2021}).
\newblock


\bibitem[\protect\citeauthoryear{Mezlini, Shapiro, Daza, Caddigan, Ramirez,
  et~al\mbox{.}}{Mezlini et~al\mbox{.}}{2021}]%
        {mezlini2021estimating}
\bibfield{author}{\bibinfo{person}{Aziz Mezlini}, \bibinfo{person}{Allison
  Shapiro}, \bibinfo{person}{Eric~J Daza}, \bibinfo{person}{Eamon Caddigan},
  \bibinfo{person}{Ernesto Ramirez}, {et~al\mbox{.}}}
  \bibinfo{year}{2021}\natexlab{}.
\newblock \showarticletitle{Estimating the Burden of Influenza on Daily
  Activity at Population Scale Using Commercial Wearable Sensors}.
\newblock \bibinfo{journal}{\emph{medRxiv}} (\bibinfo{year}{2021}).
\newblock


\bibitem[\protect\citeauthoryear{Mikolov, Sutskever, Chen, Corrado, and
  Dean}{Mikolov et~al\mbox{.}}{2013}]%
        {mikolov2013distributed}
\bibfield{author}{\bibinfo{person}{Tomas Mikolov}, \bibinfo{person}{Ilya
  Sutskever}, \bibinfo{person}{Kai Chen}, \bibinfo{person}{Greg~S Corrado},
  {and} \bibinfo{person}{Jeff Dean}.} \bibinfo{year}{2013}\natexlab{}.
\newblock \showarticletitle{Distributed representations of words and phrases
  and their compositionality}. In \bibinfo{booktitle}{\emph{Advances in neural
  information processing systems}}. \bibinfo{pages}{3111--3119}.
\newblock


\bibitem[\protect\citeauthoryear{Mishra, Wang, Metwally, Bogu, Brooks,
  et~al\mbox{.}}{Mishra et~al\mbox{.}}{2020}]%
        {mishra_pre-symptomatic_2020}
\bibfield{author}{\bibinfo{person}{Tejaswini Mishra}, \bibinfo{person}{Meng
  Wang}, \bibinfo{person}{Ahmed~A. Metwally}, \bibinfo{person}{Gireesh~K.
  Bogu}, \bibinfo{person}{Andrew~W. Brooks}, {et~al\mbox{.}}}
  \bibinfo{year}{2020}\natexlab{}.
\newblock \showarticletitle{Pre-symptomatic detection of {COVID}-19 from
  smartwatch data}.
\newblock \bibinfo{journal}{\emph{Nature Biomedical Engineering}}
  (\bibinfo{year}{2020}).
\newblock


\bibitem[\protect\citeauthoryear{Nair, Javkar, Wu, and Frias-Martinez}{Nair
  et~al\mbox{.}}{2019}]%
        {nair_understanding_2019}
\bibfield{author}{\bibinfo{person}{Suraj Nair}, \bibinfo{person}{Kiran Javkar},
  \bibinfo{person}{Jiahui Wu}, {and} \bibinfo{person}{Vanessa Frias-Martinez}.}
  \bibinfo{year}{2019}\natexlab{}.
\newblock \showarticletitle{Understanding {Cycling} {Trip} {Purpose} and
  {Route} {Choice} {Using} {GPS} {Traces} and {Open} {Data}}.
\newblock \bibinfo{journal}{\emph{IMWUT}} (\bibinfo{year}{2019}).
\newblock


\bibitem[\protect\citeauthoryear{Natarajan, Su, and Heneghan}{Natarajan
  et~al\mbox{.}}{2020}]%
        {natarajan_assessment_2020}
\bibfield{author}{\bibinfo{person}{Aravind Natarajan}, \bibinfo{person}{Hao-Wei
  Su}, {and} \bibinfo{person}{Conor Heneghan}.}
  \bibinfo{year}{2020}\natexlab{}.
\newblock \showarticletitle{Assessment of physiological signs associated with
  {COVID}-19 measured using wearable devices}.
\newblock \bibinfo{journal}{\emph{Nature Digital Medicine}}
  (\bibinfo{year}{2020}).
\newblock


\bibitem[\protect\citeauthoryear{Nestor, Hunter, Kainkaryam, Drysdale, Inglis,
  et~al\mbox{.}}{Nestor et~al\mbox{.}}{2021}]%
        {nestor2021dear}
\bibfield{author}{\bibinfo{person}{Bret Nestor}, \bibinfo{person}{Jaryd
  Hunter}, \bibinfo{person}{Raghu Kainkaryam}, \bibinfo{person}{Erik Drysdale},
  \bibinfo{person}{Jeffrey~B Inglis}, {et~al\mbox{.}}}
  \bibinfo{year}{2021}\natexlab{}.
\newblock \showarticletitle{Dear Watch, Should I Get a COVID-19 Test? Designing
  deployable machine learning for wearables}.
\newblock \bibinfo{journal}{\emph{medRxiv}} (\bibinfo{year}{2021}).
\newblock


\bibitem[\protect\citeauthoryear{Ni, Muhlstein, and McAuley}{Ni
  et~al\mbox{.}}{2019}]%
        {ni2019modeling}
\bibfield{author}{\bibinfo{person}{Jianmo Ni}, \bibinfo{person}{Larry
  Muhlstein}, {and} \bibinfo{person}{Julian McAuley}.}
  \bibinfo{year}{2019}\natexlab{}.
\newblock \showarticletitle{Modeling heart rate and activity data for
  personalized fitness recommendation}. In \bibinfo{booktitle}{\emph{WWW}}.
\newblock


\bibitem[\protect\citeauthoryear{Pyrkov, Slipensky, Barg, Kondrashin, Zhurov,
  et~al\mbox{.}}{Pyrkov et~al\mbox{.}}{2018}]%
        {pyrkov_extracting_2018}
\bibfield{author}{\bibinfo{person}{Timothy~V. Pyrkov},
  \bibinfo{person}{Konstantin Slipensky}, \bibinfo{person}{Mikhail Barg},
  \bibinfo{person}{Alexey Kondrashin}, \bibinfo{person}{Boris Zhurov},
  {et~al\mbox{.}}} \bibinfo{year}{2018}\natexlab{}.
\newblock \showarticletitle{Extracting biological age from biomedical data via
  deep learning: too much of a good thing?}
\newblock \bibinfo{journal}{\emph{Scientific Reports}} (\bibinfo{year}{2018}).
\newblock


\bibitem[\protect\citeauthoryear{Quer, Radin, Gadaleta, Baca-Motes, Ariniello,
  et~al\mbox{.}}{Quer et~al\mbox{.}}{2020}]%
        {quer_wearable_2020}
\bibfield{author}{\bibinfo{person}{Giorgio Quer}, \bibinfo{person}{Jennifer~M.
  Radin}, \bibinfo{person}{Matteo Gadaleta}, \bibinfo{person}{Katie
  Baca-Motes}, \bibinfo{person}{Lauren Ariniello}, {et~al\mbox{.}}}
  \bibinfo{year}{2020}\natexlab{}.
\newblock \showarticletitle{Wearable sensor data and self-reported symptoms for
  {COVID}-19 detection}.
\newblock \bibinfo{journal}{\emph{Nature Medicine}} (\bibinfo{year}{2020}).
\newblock


\bibitem[\protect\citeauthoryear{Saito and Rehmsmeier}{Saito and
  Rehmsmeier}{2015}]%
        {saito_precision-recall_2015}
\bibfield{author}{\bibinfo{person}{Takaya Saito} {and} \bibinfo{person}{Marc
  Rehmsmeier}.} \bibinfo{year}{2015}\natexlab{}.
\newblock \showarticletitle{The {Precision}-{Recall} {Plot} {Is} {More}
  {Informative} than the {ROC} {Plot} {When} {Evaluating} {Binary}
  {Classifiers} on {Imbalanced} {Datasets}}.
\newblock \bibinfo{journal}{\emph{PLoS ONE}} (\bibinfo{year}{2015}).
\newblock


\bibitem[\protect\citeauthoryear{Shapiro, Marinsek, Clay, Bradshaw, Ramirez,
  et~al\mbox{.}}{Shapiro et~al\mbox{.}}{2021}]%
        {shapiro2021characterizing}
\bibfield{author}{\bibinfo{person}{Allison Shapiro}, \bibinfo{person}{Nicole
  Marinsek}, \bibinfo{person}{Ieuan Clay}, \bibinfo{person}{Benjamin Bradshaw},
  \bibinfo{person}{Ernesto Ramirez}, {et~al\mbox{.}}}
  \bibinfo{year}{2021}\natexlab{}.
\newblock \showarticletitle{Characterizing COVID-19 and influenza illnesses in
  the real world via person-generated health data}.
\newblock \bibinfo{journal}{\emph{Patterns}} (\bibinfo{year}{2021}),
  \bibinfo{pages}{100188}.
\newblock


\bibitem[\protect\citeauthoryear{Shen, Voisin, Aliamiri, Avati, Hannun,
  et~al\mbox{.}}{Shen et~al\mbox{.}}{2019}]%
        {shen_ambulatory_2019}
\bibfield{author}{\bibinfo{person}{Yichen Shen}, \bibinfo{person}{Maxime
  Voisin}, \bibinfo{person}{Alireza Aliamiri}, \bibinfo{person}{Anand Avati},
  \bibinfo{person}{Awni Hannun}, {et~al\mbox{.}}}
  \bibinfo{year}{2019}\natexlab{}.
\newblock \showarticletitle{Ambulatory {Atrial} {Fibrillation} {Monitoring}
  {Using} {Wearable} {Photoplethysmography} with {Deep} {Learning}}. In
  \bibinfo{booktitle}{\emph{KDD}}.
\newblock


\bibitem[\protect\citeauthoryear{Song, Rajan, Thiagarajan, and Spanias}{Song
  et~al\mbox{.}}{2018}]%
        {song2018attend}
\bibfield{author}{\bibinfo{person}{Huan Song}, \bibinfo{person}{Deepta Rajan},
  \bibinfo{person}{Jayaraman~J Thiagarajan}, {and} \bibinfo{person}{Andreas
  Spanias}.} \bibinfo{year}{2018}\natexlab{}.
\newblock \showarticletitle{Attend and diagnose: Clinical time series analysis
  using attention models}. In \bibinfo{booktitle}{\emph{AAAI}}.
\newblock


\bibitem[\protect\citeauthoryear{Spathis, Servia-Rodriguez, Farrahi, Mascolo,
  and Rentfrow}{Spathis et~al\mbox{.}}{2019}]%
        {spathis2019sequence}
\bibfield{author}{\bibinfo{person}{Dimitris Spathis}, \bibinfo{person}{Sandra
  Servia-Rodriguez}, \bibinfo{person}{Katayoun Farrahi},
  \bibinfo{person}{Cecilia Mascolo}, {and} \bibinfo{person}{Jason Rentfrow}.}
  \bibinfo{year}{2019}\natexlab{}.
\newblock \showarticletitle{Sequence multi-task learning to forecast mental
  wellbeing from sparse self-reported data}. In
  \bibinfo{booktitle}{\emph{KDD}}.
\newblock


\bibitem[\protect\citeauthoryear{Suhara, Xu, and Pentland}{Suhara
  et~al\mbox{.}}{2017}]%
        {suhara2017deepmood}
\bibfield{author}{\bibinfo{person}{Yoshihiko Suhara}, \bibinfo{person}{Yinzhan
  Xu}, {and} \bibinfo{person}{Alex'Sandy' Pentland}.}
  \bibinfo{year}{2017}\natexlab{}.
\newblock \showarticletitle{Deepmood: Forecasting depressed mood based on
  self-reported histories via recurrent neural networks}. In
  \bibinfo{booktitle}{\emph{WWW}}.
\newblock


\bibitem[\protect\citeauthoryear{Tang, Perez-Pozuelo, Spathis, Brage, Wareham,
  et~al\mbox{.}}{Tang et~al\mbox{.}}{2021}]%
        {tang2021selfhar}
\bibfield{author}{\bibinfo{person}{Chi~Ian Tang}, \bibinfo{person}{Ignacio
  Perez-Pozuelo}, \bibinfo{person}{Dimitris Spathis}, \bibinfo{person}{Soren
  Brage}, \bibinfo{person}{Nick Wareham}, {et~al\mbox{.}}}
  \bibinfo{year}{2021}\natexlab{}.
\newblock \showarticletitle{SelfHAR: Improving Human Activity Recognition
  through Self-training with Unlabeled Data}.
\newblock \bibinfo{journal}{\emph{arXiv:2102.06073}} (\bibinfo{year}{2021}).
\newblock


\bibitem[\protect\citeauthoryear{Vaswani, Shazeer, Parmar, Uszkoreit, Jones,
  et~al\mbox{.}}{Vaswani et~al\mbox{.}}{2017}]%
        {vaswani2017attention}
\bibfield{author}{\bibinfo{person}{Ashish Vaswani}, \bibinfo{person}{Noam
  Shazeer}, \bibinfo{person}{Niki Parmar}, \bibinfo{person}{Jakob Uszkoreit},
  \bibinfo{person}{Llion Jones}, {et~al\mbox{.}}}
  \bibinfo{year}{2017}\natexlab{}.
\newblock \showarticletitle{Attention is all you need}. In
  \bibinfo{booktitle}{\emph{Advances in neural information processing
  systems}}. \bibinfo{pages}{5998--6008}.
\newblock


\bibitem[\protect\citeauthoryear{Wang, Chen, Hao, Peng, and Hu}{Wang
  et~al\mbox{.}}{2019}]%
        {wang2019deep}
\bibfield{author}{\bibinfo{person}{Jindong Wang}, \bibinfo{person}{Yiqiang
  Chen}, \bibinfo{person}{Shuji Hao}, \bibinfo{person}{Xiaohui Peng}, {and}
  \bibinfo{person}{Lisha Hu}.} \bibinfo{year}{2019}\natexlab{}.
\newblock \showarticletitle{Deep learning for sensor-based activity
  recognition: A survey}.
\newblock \bibinfo{journal}{\emph{Pattern Recognition Letters}}
  (\bibinfo{year}{2019}).
\newblock


\bibitem[\protect\citeauthoryear{Wang, Aung, Abdullah, Brian, Campbell,
  et~al\mbox{.}}{Wang et~al\mbox{.}}{2016}]%
        {wang2016crosscheck}
\bibfield{author}{\bibinfo{person}{Rui Wang}, \bibinfo{person}{Min~SH Aung},
  \bibinfo{person}{Saeed Abdullah}, \bibinfo{person}{Rachel Brian},
  \bibinfo{person}{Andrew~T Campbell}, {et~al\mbox{.}}}
  \bibinfo{year}{2016}\natexlab{}.
\newblock \showarticletitle{CrossCheck: toward passive sensing and detection of
  mental health changes in people with schizophrenia}. In
  \bibinfo{booktitle}{\emph{UbiComp}}.
\newblock


\bibitem[\protect\citeauthoryear{Wang, Chen, Chen, Li, Harari,
  et~al\mbox{.}}{Wang et~al\mbox{.}}{2014}]%
        {wang2014studentlife}
\bibfield{author}{\bibinfo{person}{Rui Wang}, \bibinfo{person}{Fanglin Chen},
  \bibinfo{person}{Zhenyu Chen}, \bibinfo{person}{Tianxing Li},
  \bibinfo{person}{Gabriella Harari}, {et~al\mbox{.}}}
  \bibinfo{year}{2014}\natexlab{}.
\newblock \showarticletitle{StudentLife: assessing mental health, academic
  performance and behavioral trends of college students using smartphones}. In
  \bibinfo{booktitle}{\emph{UbiComp}}.
\newblock


\bibitem[\protect\citeauthoryear{Wu, Huang, Roblesgranda, and Chawla}{Wu
  et~al\mbox{.}}{2020}]%
        {wu2020representation}
\bibfield{author}{\bibinfo{person}{Xian Wu}, \bibinfo{person}{Chao Huang},
  \bibinfo{person}{Pablo Roblesgranda}, {and} \bibinfo{person}{Nitesh Chawla}.}
  \bibinfo{year}{2020}\natexlab{}.
\newblock \showarticletitle{Representation learning on variable length and
  incomplete wearable-sensory time series}.
\newblock  (\bibinfo{year}{2020}).
\newblock


\bibitem[\protect\citeauthoryear{Xu, Mankoff, and Dey}{Xu
  et~al\mbox{.}}{2021}]%
        {xu2021understanding}
\bibfield{author}{\bibinfo{person}{Xuhai Xu}, \bibinfo{person}{Jennifer
  Mankoff}, {and} \bibinfo{person}{Anind~K. Dey}.}
  \bibinfo{year}{2021}\natexlab{}.
\newblock \bibinfo{title}{Understanding Practices and Needs of Researchers in
  Human State Modeling by Passive Mobile Sensing}.
\newblock
\newblock


\bibitem[\protect\citeauthoryear{Yao, Hu, Zhao, Zhang, and Abdelzaher}{Yao
  et~al\mbox{.}}{2017}]%
        {yao2017deepsense}
\bibfield{author}{\bibinfo{person}{Shuochao Yao}, \bibinfo{person}{Shaohan Hu},
  \bibinfo{person}{Yiran Zhao}, \bibinfo{person}{Aston Zhang}, {and}
  \bibinfo{person}{Tarek Abdelzaher}.} \bibinfo{year}{2017}\natexlab{}.
\newblock \showarticletitle{Deepsense: A unified deep learning framework for
  time-series mobile sensing data processing}. In
  \bibinfo{booktitle}{\emph{WWW}}.
\newblock


\bibitem[\protect\citeauthoryear{Zerveas, Jayaraman, Patel, Bhamidipaty, and
  Eickhoff}{Zerveas et~al\mbox{.}}{2021}]%
        {zerveas_transformer-based_2021}
\bibfield{author}{\bibinfo{person}{George Zerveas}, \bibinfo{person}{Srideepika
  Jayaraman}, \bibinfo{person}{Dhaval Patel}, \bibinfo{person}{Anuradha
  Bhamidipaty}, {and} \bibinfo{person}{Carsten Eickhoff}.}
  \bibinfo{year}{2021}\natexlab{}.
\newblock \showarticletitle{A {Transformer}-based {Framework} for
  {Multivariate} {Time} {Series} {Representation} {Learning}}.
\newblock In \bibinfo{booktitle}{\emph{KDD}}.
\newblock


\bibitem[\protect\citeauthoryear{Zhang, Song, Chen, Feng, Lumezanu,
  et~al\mbox{.}}{Zhang et~al\mbox{.}}{2019}]%
        {zhang_deep_2019}
\bibfield{author}{\bibinfo{person}{Chuxu Zhang}, \bibinfo{person}{Dongjin
  Song}, \bibinfo{person}{Yuncong Chen}, \bibinfo{person}{Xinyang Feng},
  \bibinfo{person}{Cristian Lumezanu}, {et~al\mbox{.}}}
  \bibinfo{year}{2019}\natexlab{}.
\newblock \showarticletitle{A {Deep} {Neural} {Network} for {Unsupervised}
  {Anomaly} {Detection} and {Diagnosis} in {Multivariate} {Time} {Series}
  {Data}}.
\newblock \bibinfo{journal}{\emph{AAAI}} (\bibinfo{date}{July}
  \bibinfo{year}{2019}).
\newblock


\bibitem[\protect\citeauthoryear{Zhang, Xu, Li, Kostakos, Hui,
  et~al\mbox{.}}{Zhang et~al\mbox{.}}{2021}]%
        {zhang_passive_2021}
\bibfield{author}{\bibinfo{person}{Yunke Zhang}, \bibinfo{person}{Fengli Xu},
  \bibinfo{person}{Tong Li}, \bibinfo{person}{Vassilis Kostakos},
  \bibinfo{person}{Pan Hui}, {et~al\mbox{.}}} \bibinfo{year}{2021}\natexlab{}.
\newblock \showarticletitle{Passive {Health} {Monitoring} {Using} {Large}
  {Scale} {Mobility} {Data}}.
\newblock \bibinfo{journal}{\emph{IMWUT}} (\bibinfo{year}{2021}).
\newblock


\end{thebibliography}

\balance
\clearpage
\renewcommand\thefigure{\thesection.\arabic{figure}} 
\renewcommand\thetable{\thetable.\arabic{table}} 
\begin{appendix}
\onecolumn
\section{Reproducability Appendix}

\setcounter{figure}{0}  
\setcounter{table}{0}  
\counterwithin{table}{section}

\subsection{Hyperparameter Tuning}
\label{sec:hyperparam}
All models in this paper were trained with a randomized hyperparameter sweep using a withheld validation set. For CNN modules, we experimented with kernel sizes as large as 63, stride sizes as large as 256, depths as deep as eight layers, and as many as 32 output channels. For transformers modules, we experimented pre-computed and fixed positional embeddings, up to twelve layers of stacked transformers, and up to nine-head attention. We also tried dropout rates between 0.0 and 0.5. In total, over five hundred model configurations were tested before setting on the final configuration of kernel sizes of 5,5,2, stride sizes of 5,3,2, output channels of 8,16,32, two transformer layers each with four heads, and dropout of 0.4.  We tried Adam learning rates from 1 to $1e-6$, and found that $5e-4$ worked best. This relatively small learning rate seemed to be important for limiting overfitting. We also conducted a hyperparameter sweep for XGBoost models, and found little low to hyperparameters, and that $\eta=1$ and a maximum depth of six worked best. Further, we experimented with window sizes ranging from three to ten days, and found that the model overfitted on both ends of this range, with best performance at seven days. 

\binnedMissing
\tableDataStats
\pretrainConcept
\tableZeroShotFeatures
\end{appendix}

\end{document}